\documentclass{article} % For LaTeX2e
\usepackage{iclr2027_conference,times}

\usepackage{amsmath,amsfonts,bm}

\def\eqref#1{equation~\ref{#1}}
\def\1{\bm{1}}

\DeclareMathAlphabet{\mathsfit}{\encodingdefault}{\sfdefault}{m}{sl}
\SetMathAlphabet{\mathsfit}{bold}{\encodingdefault}{\sfdefault}{bx}{n}

\usepackage{hyperref}
\usepackage{url}
\usepackage{wrapfig}

\usepackage{booktabs}
\usepackage[most]{tcolorbox}
\definecolor{FigureBlue}{HTML}{427EB4}

\newtcolorbox{findingbox}{
  enhanced,
  colback=white,
  colframe=FigureBlue!45!white,
  boxrule=0.6pt,
  arc=2pt,
  borderline west={5pt}{0pt}{FigureBlue},
  left=12pt, right=9pt, top=7pt, bottom=7pt,
  before skip=6pt, after skip=6pt
}
\usepackage{multirow}
\usepackage{graphicx}
\usepackage{xcolor}
\usepackage{tikz}
\usetikzlibrary{calc}
\usepackage{pifont}
\usepackage{amsmath,amssymb}
\usepackage{enumitem}
\usepackage{makecell}
\definecolor{checkgreen}{RGB}{0,145,70}
\definecolor{crossred}{RGB}{205,35,35}
\definecolor{partialorange}{RGB}{230,125,20}
\newcommand{\cmark}{\textcolor{checkgreen}{\ding{51}}}
\newcommand{\xmark}{\textcolor{crossred}{\ding{55}}}
\newcommand{\pmark}{\textcolor{partialorange}{$\triangle$}}

\newcommand{\na}{--}
\title{VibeTraining-Bench: Data–Trainer–Harness Co-Design for Iterative System Improvement}

\iclrfinalcopy
\author{\textbf{Ruotian Peng\textsuperscript{1,2,3,*}\quad
Yuanyi Ren\textsuperscript{3,*}\quad
Shuaiqi Liu\textsuperscript{3}\quad
Xin Guo\textsuperscript{3}}\quad
\textbf{Fang Zhu\textsuperscript{4}\quad
Yangyi Huang\textsuperscript{5}} \\
\textbf{Yue Wu\textsuperscript{3}\quad Zheng Wang\textsuperscript{3}\quad
Yandong Wen\textsuperscript{2,\textdagger}\quad
Zhihang Fu\textsuperscript{3,\textdagger}}
 \vspace{3pt}
 \\
 \textsuperscript{1}Zhejiang University~~~
 \textsuperscript{2}Westlake University~~~
 \textsuperscript{3}Alibaba Token Hub, Alibaba Group~~~ \\
 \textsuperscript{4}South China University of Technology~~~
 \textsuperscript{5}The Chinese University of Hong Kong
 \\[0.5mm]
$^*$Equal contribution~~~~$^\dagger$Corresponding author
}

\newtcblisting{researchpromptbox}{
  enhanced,
  breakable,
  listing only,
  colback=white,
  colframe=FigureBlue,
  colbacktitle=FigureBlue,
  coltitle=white,
  title={Research Agent Workflow Prompt},
  fonttitle=\bfseries\small,
  boxrule=0.6pt,
  arc=2pt,
  left=8pt,
  right=8pt,
  top=7pt,
  bottom=7pt,
  listing options={
    basicstyle=\ttfamily\footnotesize,
    breaklines=true,
    breakatwhitespace=true,
    columns=fullflexible,
    keepspaces=true,
    showstringspaces=false
  }
}

\usepackage{tabularx,array,placeins}
\definecolor{pendinggray}{RGB}{94,102,113}
\definecolor{pendingbg}{RGB}{246,247,249}

\newcolumntype{Y}{>{\raggedright\arraybackslash}X}

\begin{document}

\maketitle

\begin{abstract}
% A key challenge in recursive self-improvement is enabling research agents to iteratively improve an agent system as a whole rather than optimize a predefined component. Agent system performance depends on training data, the trainer, and the inference harness, and the most effective changes may shift across components as the system evolves. Existing benchmarks primarily focus on individual components or training-side optimization, leaving research agents' ability to select and coordinate changes across training and inference largely unexplored. We refer to this capability as \emph{cross-component co-design}.
% 这里的逻辑： RSI 的核心为 Imp Mechanism，即诊断问题与提出改进的能力 +  Imp Mechanism 的对象为多组件耦合的复杂系统 -> agent需要具备判断改哪里，如何协同的核心能力 -> 现有 benchmark 把 Imp 的可见域缩到单组件，无法验证这个能力
%%%%%%%%% RYY改 %%%%%%%%%
% Recursive self-improvement (RSI), increasingly studied as a path to autonomous AI development, requires agents to diagnose bottlenecks and devise effective strategies to improve agent performance. Addressing these bottlenecks involves decisions about training data, the trainer, and the inference harness, with each component’s contribution depending on the configuration of the others. Agents must therefore determine where to intervene and how to coordinate modifications across training and inference. We refer to this capability as \emph{cross-component co-design}. Existing benchmarks primarily evaluate optimization within individual components or training pipelines, leaving this capability insufficiently assessed.
Recursive self-improvement (RSI) requires agents to diagnose system bottlenecks and coordinate improvements across training and inference. We refer to the agents’ capability to jointly optimize training data, the trainer, and the inference harness as \emph{cross-component co-design}.
%%%%%%%%%%%%%%%%%%%%%%%%%
We introduce \textbf{VibeTraining-Bench}, a benchmark for evaluating this capability across tasks spanning terminal operation, software engineering, multi-turn tool use, and mathematical reasoning. Within each task, research agents start from an identical initial agent system under a fixed resource budget. 
Guided by validation feedback, they iteratively refine data, trainers, and inference harnesses, then select and submit a final agent system for evaluation on the test set. 
%Guided by validation feedback and execution trajectories, they iteratively modify data, trainer, and inference harnesses, ultimately producing a final agent system evaluated on test tasks. 
To prevent reward hacking, we enforce three safeguards: test data isolation, separate research and grading containers, and network restrictions for research agents. 
Our evaluation results show that frontier research agents substantially improve target agent systems. For example, on SWE-bench Verified, GPT-6-Astra raises the target agent system's performance from 32.9\% to 53.7\%, comparable to the human-developed instruct model's 55.4\%. 
%However, research agents explore the editable components unevenly, and validation gains do not consistently translate into improvements on test tasks. 
However, there is still room to improve their ability to co-design and iteratively refine data, trainer, and harness to achieve cumulative performance gains. 
We hope that VibeTraining-Bench will serve as a unified platform for advancing RSI, deepening our understanding of cross-component optimization for agent systems.
Code and data are available at \url{https://github.com/D2I-ai/VibeTraining-Bench}.
\end{abstract}

\section{Introduction}
Recursive self-improvement (RSI), increasingly studied as a path toward autonomous AI development, requires agents to diagnose performance bottlenecks and propose and validate modifications to an agent system~\citep{202608.0051,yang2026frontis}. These modifications may involve training data, the trainer, and the inference harness, with their effectiveness depending on how these components are configured together~\citep{ursekar2026harnessopt}. For example, the harness controls what an agent observes and can do, steering its exploration and thus the trajectories the model learns from; conversely, the model's capabilities determine what the harness must provide~\citep{chen2026evotrainer,chen2026co}. Agents must therefore trace failures to their sources and coordinate changes across training and inference. We refer to this capability as \emph{cross-component co-design}.
%%%%%%%%%%%%%%%%%%%%%%%%%

% Existing benchmarks do not directly evaluate cross-component co-design of an evolving target agent system. 
Assessing cross-component co-design thus calls for settings in which agents themselves determine which components to modify and how to coordinate these modifications as the system evolves. To the best of our knowledge, few benchmarks directly assess this capability.
MLE-Bench and RE-Bench evaluate research agents on diverse machine learning engineering and research tasks, focusing on task-specific research outcomes. More directly related RSI benchmarks often evaluate single-component improvement. RSIBench-Data~\citep{meng2026rsibench}, AI4AI-Bench~\citep{chi2026ai4ai}, and HarnessOpt-Bench~\citep{ursekar2026harnessopt} focus on data strategies, training algorithms, and inference harnesses, respectively; each MLS-Bench task specifies one ML component to improve~\citep{lyu2026mls}. These benchmarks evaluate improvements within predefined
component scopes. PostTrainBench~\citep{rank2026posttrainbench} and Agent$^2$ RL-Bench~\citep{chen2026agent} broaden this scope to data construction and post-training procedures. Simply broadening the editable scope also raises evaluation risks: PostTrainBench reports reward hacking, including research agents training on test data~\citep{rank2026posttrainbench}. Even these multi-component benchmarks remain focused on
training-side optimization, leaving the iterative selection
and coordination of changes across training and inference
within the same evolving target system unexplored.

We introduce \textsc{VibeTraining-Bench} to evaluate cross-component co-design through five tasks spanning terminal operation, software engineering, multi-turn tool use, and mathematical reasoning.
In each task, an agent receives an initial agent system comprising a base model, open-source training data, a trainer with algorithm code and initial scripts, and an inference harness. The agent must improve this system within a fixed resource budget, and the agent it finally submits is then scored on hidden test instances. 
% The benchmark design addresses both requirements of cross-component co-design. 
% First, the agent can evaluate candidate systems on validation instances throughout a run and receives scores together with execution trajectories. This feedback allows it to trace failures to their sources, identify the components responsible, and assess the effect of each revision in subsequent iterations. 
% Second, rather than prescribing which components to optimize, the benchmark grants the agent access to multiple components, allowing it to act on its diagnosis and coordinate changes across them. Training data and the trainer are editable in every task, and so is the inference harness whenever the task's evaluation does not prescribe a particular harness.
During a run, validation scores and execution trajectories help the agent trace failures to their sources and assess successive revisions. The benchmark also allows the agent to choose where to intervene and coordinate changes across components. Training data and the trainer are editable in every task, as is the inference harness unless the task's evaluation prescribes a particular harness.
These design choices, however, also widen the opportunities for reward hacking, so \textsc{VibeTraining-Bench} further enforces three safeguards, namely data isolation, container isolation, and network restrictions. Specifically, test data remain inaccessible throughout the research phase; research and grading run in separate containers, preventing research agents from modifying grader code or evaluation data; and research environments strictly restrict external network access.
%%%%%%%%%%%%%%%%%%%%%%%%%

Our evaluation results show that frontier research agents substantially improve target agent systems. 
%, although no single configuration leads across all tasks. 
For example, on SWE-bench Verified, GPT-6-Astra coordinates training and harness changes, raising the target agent system's performance from 32.9\% to 53.7\%, comparable to the human-developed instruct model's 55.4\%. However, research agents explore the improvable system components unevenly. Analysis of research trajectories shows that agents primarily focus on lightweight refinements, such as data filtering and training hyperparameter tuning. They less frequently explore changes to the underlying training algorithms or coordinated modifications across training data, the trainer, and the inference harness. 
There is still room for improvement in research agents' cross-component co-design capability and iterative refinement toward achieving cumulative performance gains. 
%Moreover, validation gains do not consistently translate into improvements on test tasks.

In summary, we make the following contributions:
\begin{itemize}[leftmargin=*,nosep]
\setlength\itemsep{0.43em}

\item 
We introduce \textsc{VibeTraining-Bench}, a benchmark spanning diverse tasks for evaluating cross-component co-design of target agent systems. Its improvement scope encompasses training data, the trainer, and the inference harness.

\item 
We develop a reliable execution platform to guarantee evaluation integrity. To prevent research agents from reward hacking, our platform implements three levels of safeguards by isolating test data, separating research and grading containers, and restricting external network access.

\item 
We evaluate frontier research agents through test performance, research trajectories, and component ablations. Research agents substantially improve target agent systems, in some cases approaching the performance of the system with human post-training. 
Our experimental analysis reveals that research agents built upon frontier models and harnesses still exhibit limitations on these tasks, indicating substantial room for further improvement. 
%Our analysis also reveals uneven component exploration and validation gains that do not consistently translate into test improvements.
\end{itemize}
\section{Related Work}
\begin{table*}[t]
\centering
\caption{Protocol comparison of benchmarks for recursive self-improvement.
\textit{Legend:} \cmark{} = enforced; \pmark{} = partial or insufficiently documented; \xmark{} = unsupported; \na{} = not applicable.
Trajectory feedback requires task-level traces or diagnostics, and iterative revision requires repeated propose--evaluate--revise cycles.
Joint requires coordinated changes to training data, the trainer, and the inference harness within one task.
Dev/test isolation keeps the final test inaccessible during development; train/eval isolation separates training, serving, and evaluation services.}
\vspace{6pt}
\label{tab:benchmark-comparison}
% \scriptsize
\footnotesize
\setlength{\tabcolsep}{1.5pt}
\renewcommand{\arraystretch}{1.22}
% \resizebox{\textwidth}{!}{%
% \small
\begin{tabular}{lcccccccc}
\toprule
\multirow{2}{*}[-4pt]{\textbf{Benchmark}} &
\multicolumn{2}{c}{\textbf{Research Loop}} &
\multicolumn{2}{c}{\textbf{Editable Training}} &
\multicolumn{2}{c}{\textbf{System Co-Design}} &
\multicolumn{2}{c}{\textbf{Controlled Infrastructure}} \\
\cmidrule(lr){2-3}\cmidrule(lr){4-5}\cmidrule(lr){6-7}\cmidrule(lr){8-9}
& \makecell{Trajectory\\Feedback} & \makecell{Iterative\\Revision} & Data & Trainer & Harness & Joint &\makecell{Dev/Test\\Isolation} & \makecell{Train/Eval\\Isolation} \\
\midrule
MLS-Bench

  & \pmark & \cmark & \xmark & \cmark & \xmark & \xmark & \pmark & \xmark \\
PostTrainBench

  & \cmark & \xmark & \cmark & \cmark & \xmark & \xmark & \xmark & \xmark \\
Curation-Bench
 
  & \cmark & \cmark & \cmark & \xmark & \xmark & \xmark & \xmark & \pmark \\
RSIBench-Data

  & \cmark & \cmark & \cmark & \xmark & \xmark & \xmark & \xmark & \cmark \\
Agent$^{2}$ RL-Bench

  & \xmark & \cmark & \cmark & \cmark & \xmark & \xmark & \xmark & \cmark  \\
AI4AI-Bench

  & \pmark & \cmark & \xmark & \cmark & \xmark & \xmark & \cmark & \cmark  \\
HarnessOpt-Bench

  & \cmark & \cmark & \xmark & \xmark & \cmark & \xmark & \cmark & \na  \\
Evo-Bench
 
  & \cmark & \cmark & \xmark & \xmark & \cmark & \xmark & \cmark & \na  \\
\midrule
\textbf{VibeTraining-Bench}

  & \cmark & \cmark & \cmark & \cmark & \cmark & \cmark & \cmark & \cmark \\
\bottomrule
\end{tabular}%
% }
% \vspace{2pt}
\end{table*}

\paragraph{Self-evolving language models and agents.}
Self-improving agent systems expose several non-exclusive intervention surfaces: training data, trainer, and inference-time harnesses.
On the data front, prior work improves models through automated experience generation, student-feedback loops, and self-synthesized trajectories~\citep{zweiger2026self, huang2026r, zhao2026absolute}.
On the trainer front, recent methods leverage LLMs to discover novel optimization objectives and adaptive reinforcement learning strategies~\citep{lu2024discovering,fang2026llmzero}.
On the inference front, self-evolving agent frameworks optimize inference harnesses by tuning prompts, tool interfaces, and runtime control flow, thereby enhancing performance under frozen model weights \citep{yin2025godel, zhang2026darwin}.
More recently, work has begun to explore multi-component coupling. For example, \cite{chen2026harnessx} reuse inference rollouts for training, while \cite{chen2026evotrainer} co-evolve harness and model weights; nevertheless, these interactions follow static, predefined pathways.
Taken together, these efforts reflect an emerging shift toward coupled system-level optimization. \textsc{VibeTraining-Bench} advances this paradigm to an open-ended setting, evaluating whether research agents can autonomously decide where and how to coordinate improvements across data, trainer, and inference harnesses without a prescribed recipe.

% \vspace{-5mm}
\paragraph{Benchmarks for autonomous ML research and recursive self-improvement.}
% Existing benchmarks evaluate autonomous ML research across engineering, experimentation, replication, and method discovery \citep{huang2024mlagentbench,chan2025mle,qiang2025mledojo,wijk2024re,starace2025paperbench,lyu2026mlsbenchholisticrigorousassessment}.
% Recent benchmarks for AI improvement vary in how much of the improvement loop is placed under researcher-agent control.
% Curation-Bench and RSIBench-Data focus on data-centric research \citep{kang2026curationbench,meng2026rsibench}; AI4AI-Bench focuses on learning-algorithm design \citep{chi2026ai4ai}; and HarnessOpt-Bench and Evo-Bench optimize inference-time harnesses while keeping the underlying policy model fixed \citep{ursekar2026harnessopt,huang2026evobench}.
% PostTrainBench and Agent$^2$ RL-Bench broaden this scope to data construction and training-procedure design, but focus on post-training rather than joint optimization of the deployed inference-time harness \citep{rank2026posttrainbench,chen2026agent}.
% \textsc{VibeTraining-Bench} instead places training data, trainer, and inference-time harnesses jointly under researcher-agent control, while using disjoint development and final-evaluation data and separate services to prevent the final benchmark evaluation from becoming an optimization signal.
Existing benchmarks evaluate autonomous ML research across diverse axes, including engineering, experimentation, replication, and algorithmic discovery~\citep{huang2023mlagentbench, wijk2024re, chan2025mle, starace2025paperbench, lyu2026mls}. Within the context of recursive self-improvement (RSI), existing benchmarks differ primarily in the scope of system components exposed to the agent. To ensure attribution clarity, most prior works isolate a single component: Curation-Bench~\citep{kang2026can} and RSIBench-Data~\citep{meng2026rsibench} focus on data curation; AI4AI-Bench~\citep{chi2026ai4ai} targets training algorithms, while HarnessOpt-Bench~\citep{ursekar2026harnessopt} and Evo-Bench~\citep{huang2026evo} optimize inference scaffolding around fixed model weights. PostTrainBench and Agent$^2$ RL-Bench broaden this scope to data construction and training-procedure design, yet keep the deployed inference harness outside the optimization loop~\citep{rank2026posttrainbench,chen2026agent}. 
Table~\ref{tab:benchmark-comparison} summarizes these protocol differences across the research loop, editable components, and evaluation isolation.
\textsc{VibeTraining-Bench} bridges this gap by enabling cross-component co-design, placing training data, the trainer, and the inference harness jointly under research agent control. Crucially, because expanding the editable space naturally heightens the risk of reward hacking~\citep{rank2026posttrainbench}, our protocol strictly isolates the research agent's workspace from the grading service, preventing test-set leakage during iterative improvement.

\vspace{-2mm}
\section{VibeTraining-Bench}
\vspace{-2mm}
\begin{figure*}[t!]
  % \captionsetup{font=small, labelfont=small}
  \centering
  \setlength{\abovecaptionskip}{4pt}
  \setlength{\belowcaptionskip}{-7pt}
  \includegraphics[width=0.99\textwidth]{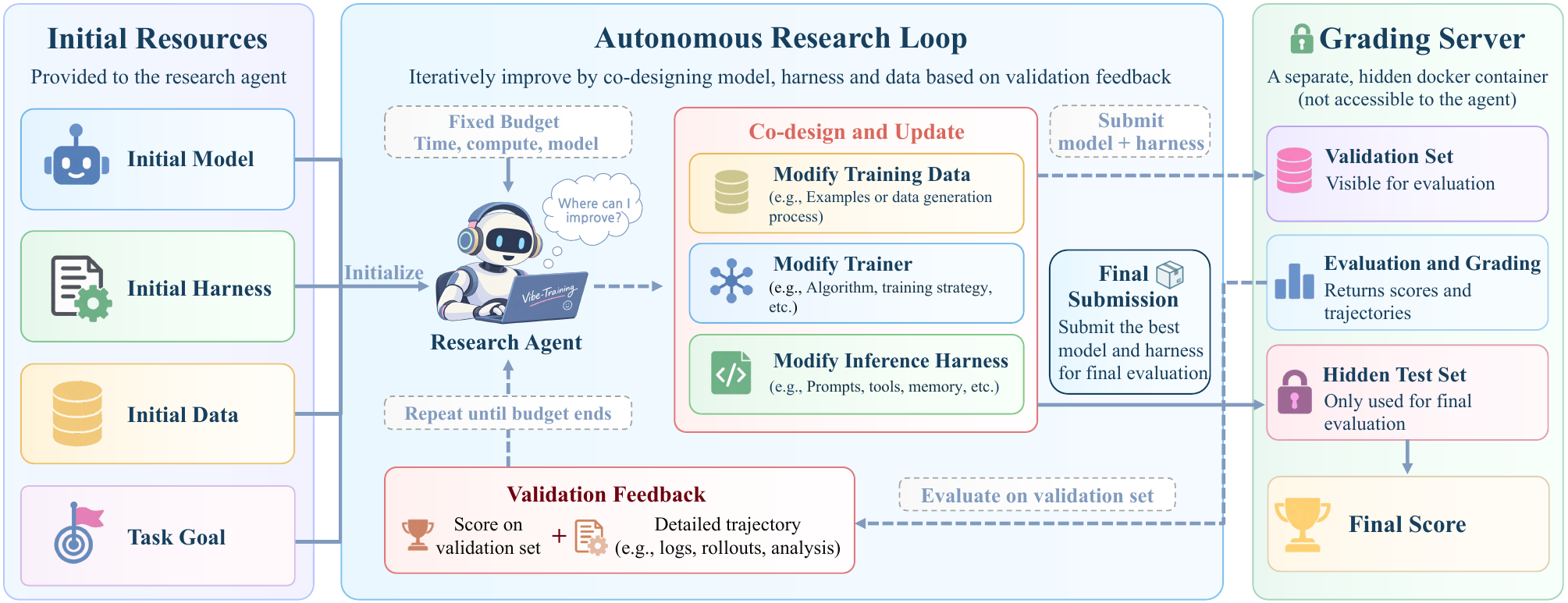}
\caption{\textbf{Overview of \textsc{VibeTraining-Bench}.} The benchmark assesses a research agent's capacity for systemic co-design across three interdependent components: training data, the trainer, and the inference harness. \textbf{(Left)} Research agents initialize from domain-specific data sources and task-specific resource budgets shared across research-agent configurations. \textbf{(Center)} Within the autonomous loop, research agents iteratively orchestrate component modifications using validation signals until budget exhaustion. \textbf{(Right)} Strict architectural decoupling isolates the optimization workspace from the grading server to prevent reward hacking. Only the visible validation set provides feedback; the hidden test set remains inaccessible until independent test evaluation of the selected system.}
\label{fig:teaser}
\end{figure*}
% \subsection{Overview and Task Formulation}
% \label{sec:task-formulation}

\subsection{Overview and Task Formulation}
\label{sec:task-formulation}

VibeTraining-Bench evaluates whether a research agent can improve a target agent system by iteratively selecting and coordinating changes to its training data, trainer, and, where editable, inference harness. Each task defines the permitted components and a fixed resource budget. Within this space, the agent uses validation feedback to decide what to modify next and which candidate target agent system to retain. 
Its objective is to select an agent system that outperforms the initial agent system on hidden test problems outside the research loop.

We formalize this process as follows:
Each task supplies a base model $M_0$, initial training data an trainer, an inference harness $H_0$, and a resource budget $\mathcal{B}$. The initial target agent system is $P_0=(M_0,H_0)$, and the research agent receives its validation score and execution trajectories before making any modifications. At research iteration $t$, the research agent may modify the training data $D_t$, trainer $T_t$, or harness $H_t$ within the task's editable space. The trainer includes the training algorithm and its configuration. The candidate target agent system is $P_t=(M_t,H_t)$, where $M_t$ may be newly trained, constructed by merging existing models, or reused from an earlier iteration. Validation returns a score and per-task execution trajectories on the requested validation tasks, which guide subsequent iterations within the resource budget.
Finally, the research agent selects one candidate target agent system as $P^\star$ using information available during research. An isolated grader evaluates this system on hidden test problems disjoint from the validation problems. We measure improvement over the initial target agent system as
\begin{equation}
\Delta_{\mathrm{test}}
=
\operatorname{Eval}_{\mathrm{test}}(P^\star)
-
\operatorname{Eval}_{\mathrm{test}}(P_0).
\end{equation}
The test result measures the selected target agent system's improvement, while the recorded attempts show which components the research agent chose to modify.
\subsection{Benchmark Suite}
\label{sec:benchmark-suite}

\noindent\textbf{Task scope.}
We instantiate the proposed cross-component co-design formulation across five target environments within VibeTraining-Bench, spanning terminal operation, software engineering, multi-turn tool use, and mathematical reasoning. As summarized in Table~\ref{tab:benchmark-suite}, these benchmarks present heterogeneous interaction paradigms under a shared iterative improvement framework. Specifically, Terminal-Bench~2.1 and SWE-bench Verified require long-horizon execution grounded in operating systems and codebases; $\tau^3$-Bench and Toolathlon demand stateful tool utilization over multi-turn dialogues; and AIME evaluates mathematical reasoning without external tool execution. Consequently, this diversity in task formulations allows us to examine whether research agents adapt their orchestration strategies across heterogeneous operational contexts.

\begin{table*}[t]
\centering
\caption{
Task composition of VibeTraining-Bench. The editable space indicates whether
the research agent may modify training data ($D$), the trainer ($T$), and the
inference harness ($H$).
}
\label{tab:benchmark-suite}

\vspace{6pt}
\small
\setlength{\tabcolsep}{4.4pt}
\renewcommand{\arraystretch}{1.03}

% Make X columns vertically centered
\renewcommand{\tabularxcolumn}[1]{m{#1}}

\begin{tabularx}{\linewidth}{
    @{}
    >{\raggedright\arraybackslash}m{0.13\linewidth}
    >{\raggedright\arraybackslash}m{0.115\linewidth}
    >{\raggedright\arraybackslash}m{0.120\linewidth}
    >{\raggedright\arraybackslash}m{0.105\linewidth}
    >{\raggedright\arraybackslash}X
    >{\centering\arraybackslash}m{0.065\linewidth}
    @{}
}
\toprule

\textbf{Capability} &
\textbf{Benchmark} &
\makecell[l]{\textbf{Initial}\\\textbf{Harness}} &
\makecell[l]{\textbf{Editable}\\\textbf{Space}} &
\textbf{Validation / Hidden Test} &
\textbf{Budget} \\

\midrule

Terminal\newline operation &
Terminal\newline -Bench~2.1 &
Terminus 2 &
$\{D,T,H\}$ &
OpenThoughts-TBLite /\newline Terminal-Bench 2.1 &
24\,h \\

Software\newline engineering &
SWE-bench\newline Verified &
mini-SWE-\newline agent &
$\{D,T,H\}$ &
Validation subset / remaining test tasks &
24\,h \\

Multi-turn\newline tool use &
$\tau^3$-Bench &
Benchmark-\newline provided &
$\{D,T\}$ &
Validation subset / remaining test tasks &
16\,h \\

Multi-turn\newline tool use &
Toolathlon &
Benchmark-\newline provided &
$\{D,T\}$ &
ToolGym / Toolathlon &
16\,h \\

Mathematical\newline reasoning &
AIME &
None &
$\{D,T\}$ &
AIME 2025 / AIME 2026 &
16\,h \\

\bottomrule
\end{tabularx}
\end{table*}

% \noindent\textbf{Editable components.}
% We systematically define the editable space of each target agent system over three components: the training data, the trainer, and the inference harness. To prevent reward hacking via external retrieval while ensuring a highly expressive optimization space, all experiments are conducted in a strictly air-gapped environment. Within this isolated sandbox, the research agent is granted unrestricted access to pre-provisioned data repositories alongside the complete source code of the training framework. This universal configuration empowers the agent to autonomously modify training algorithms, hyperparameters, and data curation pipelines across all tasks. 
% The structural role of the inference harness, however, varies depending on the benchmark. For Terminal-Bench~2.1 and SWE-bench, the harness encompasses complex control flows—including iterative execution loops, command-line invocations, and memory management—functioning as integral problem-solving capabilities rather than evaluation infrastructure. We therefore expose it to support comprehensive cross-component co-design. 
% For $\tau^3$-Bench and Toolathlon, the benchmark-provided execution adapter remains fixed. Their serving path reads exposed settings from the submitted model's generation configuration, so H is limited to effective decoding changes in these tasks. AIME fixes the evaluation interface and decoding settings, leaving D and T editable.

\noindent\textbf{Editable components.}
We systematically define the editable space of each target agent system over three components: the training data, the trainer, and the inference harness. Across all tasks, research agents can access the pre-provisioned data repositories and training framework code to modify training data, algorithms, and hyperparameters. Whether the inference harness is editable depends on the task. 
For Terminal-Bench~2.1 and SWE-bench Verified, the inference harness controls how the  target agent system interacts with terminal and codebase environments, including execution loops, command-line invocations, and memory management. We therefore make the harness editable in these two tasks to support comprehensive cross-component co-design of training data, the trainer, and inference-time behavior within the same target agent system.
For $\tau^3$-Bench and Toolathlon, the benchmark-provided execution adapters remain fixed. AIME also fixes its evaluation interface and decoding settings. These three tasks therefore expose training data and the trainer, but not the inference harness, for modification.
\subsection{Validation and Isolated Grading}
\label{sec:research-protocol}

% \noindent\textbf{Grading boundary.}
% To ensure the integrity of the iterative improvement process, VibeTraining-Bench enforces architectural isolation between the research agent's workspace and the evaluation infrastructure. The research agent operates exclusively within an air-gapped sandbox bounded by the task-specific editable space (detailed in Section~\ref{sec:benchmark-suite}). Within this boundary, the agent autonomously initiates validation requests through a controlled communication interface, retaining the flexibility to freely select specific subsets of the validation partition for targeted diagnosis. Upon receiving the updated target agent system $S_t$, all assessment procedures are executed by an independent grading service, which retains exclusive control over the task environments, verifiers, and scoring protocols. Crucially, while the agent dictates the scope of its validation queries, the grading service governs the permissible disclosure content to prevent reward hacking: it returns structured feedback $f_t$ (e.g., per-sample execution trajectories restricted to the requested subsets) while systematically withholding any signals derived from the held-out test set, thereby precluding the agent from manipulating evaluation criteria or probing the underlying test distribution.

\noindent\textbf{Grading boundary.}
To prevent reward hacking, VibeTraining-Bench enforces three safeguards: data isolation, container isolation, and network restrictions. Test tasks and test-derived feedback remain inaccessible throughout the research loop. The research agent's workspace and the grading service run in separate containers, preventing the research agent from modifying grader code or evaluation data. External network access is blocked in the research sandbox.
Research agents request validation through a controlled interface. For each request, the independent grader evaluates the candidate target agent system $P_t$, whose modifications are restricted to the task-permitted components described in Section~\ref{sec:benchmark-suite}. It returns feedback $f_t$ consisting of the validation score and per-task execution trajectories for the validation tasks. The research agent can use this feedback to guide subsequent modifications.
\medskip

% \noindent\textbf{Termination and final assessment.}
% Upon exhaustion of the predefined resource budget, the iterative improvement phase concludes, and the research agent shall execute the final nomination protocol. Specifically, the agent is required to deliver a single optimal target agent system $S^\star$ as its definitive submission. Following this delivery, the final evaluation pipeline is initiated: the aforementioned independent grading service evaluates $S^\star$ exactly once on the evaluator-private held-out test partition. Unlike the iterative validation phase, this terminal assessment yields no further feedback to the agent. We report the benchmark-native test score of the delivered system $S^\star$ alongside its absolute improvement over the initial baseline system $S_0$. The task-specific validation and test partitions are summarized in Table~\ref{tab:benchmark-suite}, with exact task identifiers and construction details provided in Appendix~\ref{app:task-splits}.

\noindent\textbf{Termination and final assessment.}
When the research phase ends, the research agent selects one candidate target agent system as $P^\star$ using the information available during research. It submits $P^\star$ to the independent grading service, which evaluates it exactly once on the test partition. Unlike iterative validation, this final assessment provides no feedback to the research agent for further revision. We report the benchmark test score of $P^\star$. Table~\ref{tab:benchmark-suite} summarizes the validation and test tasks, with additional partition details in Appendix~\ref{app:task-splits}.
% To ensure the integrity of the iterative improvement process, VibeTraining-Bench enforces architectural isolation between the research agent's workspace and the evaluation infrastructure. The research agent operates exclusively within an air-gapped sandbox bounded by the task-specific editable space (detailed in Section~\ref{sec:benchmark-suite}), interacting with the external grading service through a controlled communication interface to submit the updated target agent system $S_t$ for validation. All assessment procedures are executed by this independent grading service, which retains exclusive control over the task environments, verifiers, and scoring protocols. Crucially, to enable targeted diagnosis while preventing reward hacking, the grading service governs the permissible disclosure content: it returns structured feedback $f_t$ (e.g., per-sample execution trajectories from the validation partition) while systematically withholding any signals derived from the held-out test set, thereby precluding the agent from manipulating evaluation criteria or probing the underlying test distribution.

% -------

\FloatBarrier
\section{Experiments}
\label{sec:current-experiments}

\subsection{Setup}

We evaluate six research-agent configurations, each pairing an LLM with a coding framework. GPT-5.6-Sol and GPT-6-Astra use Codex at xhigh reasoning effort; Kimi-K3, Qwen-3.8-Max, DeepSeek-V4-Pro, and GLM-5.2 use Claude Code. Each research-agent run has access to eight NVIDIA A100 GPUs. Within each task, all configurations start from Qwen3.5-9B-Base and the same initial training data (Appendix~\ref{app:training-data}), and share the target-task interface, editable components, and task-specific resource budget (Table~\ref{tab:benchmark-suite}).
Each task provides validation tasks for iterative research and test tasks for final evaluation. Terminal-Bench~2.1 uses OpenThoughts-TBLite for validation, Toolathlon uses ToolGym, and AIME uses AIME~2025. SWE-bench Verified and $\tau^3$-Bench use disjoint validation and test partitions of their respective benchmarks.  The isolated grader evaluates the system on the hidden test using the task-specific scoring and decoding settings. We report its benchmark-native test score. More details are provided in Appendix~\ref{app:extended-figures}.

\vspace{-1mm}
\begin{table}[!htbp]
% \caption{Test performance (\%) of the selected target agent systems. Hours denotes cumulative elapsed research time across the five task runs.}
\caption{Test performance (\%) of the selected agent systems. Initial system uses Qwen3.5-9B-Base before agent-led optimization. Human-tuned baseline uses Qwen3.5-9B-Instruct, a human-developed instruct model. Subscripts indicate the harness used by each research agent; CC denotes Claude Code. Avg. denotes the mean score across the five tasks. Bold marks the highest score among the six research agents in each performance column, excluding the two baselines. Hours reports the sum of elapsed research time across the five task runs.}
\label{tab:current-results}
\centering
\vspace{6pt}
\small
\renewcommand{\arraystretch}{1.15}
\setlength{\tabcolsep}{6pt}

\begin{tabular}{lccccccc}
\toprule
\textbf{Research agent}
& \textbf{Avg.}
& \makecell{\textbf{SWE-bench}\\\textbf{Verified}}
& \makecell{\textbf{Terminal-}\\\textbf{Bench 2.1}}
& \makecell{\textbf{$\tau^3$-}\\\textbf{Bench}}
& \makecell{\textbf{Toolathlon}}
& \makecell{\textbf{AIME}\\\textbf{2026}}
& \textbf{Hours} \\
\midrule

Initial system
& 22.7
& 32.9
& 8.9
& 20.3
& 4.5
& 46.7 
& --\\

% Qwen3.5-9B-Instruct
Human-tuned baseline
& 47.3
& 55.4
& 20.2
& 64.1
& 21.6
& 75.0 
& --\\

\midrule
GPT-6-Astra\textsubscript{ +Codex}
& 37.1
& \textbf{53.7}
& 13.5
& 49.7
& 7.9
& 60.8 
& 85.9  \\

GPT-5.6-Sol\textsubscript{ +Codex}
& 31.0
& 42.9
& 8.9
& 42.1
& 10.2
& 50.8 
& 64.0\\

Kimi-K3\textsubscript{ +CC}
& \textbf{38.3}
& 51.7
& \textbf{22.4}
& \textbf{52.1}
& 10.2
& 55.0 &73.0\\

GLM-5.2\textsubscript{ +CC}
& 31.4
& 40.6
& 10.1
& 36.2
& 6.8
& \textbf{63.3}
& 56.9 \\

Qwen-3.8-Max\textsubscript{ +CC}
& 35.6
& 48.0
& 17.9
& 47.4
& 9.1
& 55.8
& 81.8 \\

DeepSeek-V4-Pro\textsubscript{ +CC}
& 34.2
& 41.4
& 13.5
& 47.9
& \textbf{12.5}
& 55.8
& 87.0 \\

\bottomrule
\end{tabular}
\end{table}

% \vspace{-2mm}
\subsection{Main Results}
\label{subsec:main-results}
% \vspace{-2mm}

\begin{wrapfigure}{r}{0.5\textwidth}
    \vspace{-1.8\baselineskip}
    \centering
    \includegraphics[width=\linewidth]
{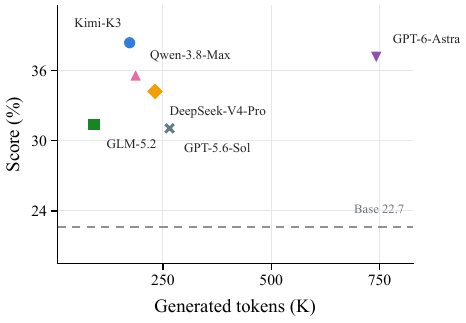}
    \caption{Mean generated tokens and macro-averaged final performance across five benchmarks. Dashed line marks the initial system score.}
    \label{fig:output-token-efficiency}
    \vspace{-0.5\baselineskip}
\end{wrapfigure}
Table~\ref{tab:current-results} reports test scores for the selected target agent systems, each comprising a selected model checkpoint and a task-specific inference harness.  The average across the five tasks is 22.7\% for the initial target agent systems and ranges from 31.1\% to 38.3\% across the six research-agent configurations, with Kimi-K3 highest. The largest gain over the initial target agent system occurs on $\tau^3$-Bench, where Kimi-K3 raises the score from 20.3\% to 52.1\%. On SWE-bench Verified, GPT-6-Astra raises the target agent system's performance from 32.9\% to 53.7\%, comparable to the human-developed instruct model's 55.4\%. On Terminal-Bench~2.1, Kimi-K3 raises the score from 8.9\% to 22.4\%, while GPT-5.6-Sol remains at 8.9\%. These results show that large gains are possible, but improvement is not guaranteed in every run.

The leader changes across tasks. GPT-6-Astra scores highest on SWE-bench Verified (53.7\%), Kimi-K3 on Terminal-Bench~2.1 (22.4\%) and $\tau^3$-Bench (52.1\%), DeepSeek-V4-Pro on Toolathlon (12.5\%), and GLM-5.2 on AIME~2026 (63.3\%). Among the four configurations using Claude Code, the gap between the highest and lowest scores ranges from 5.7 percentage points on Toolathlon to 15.9 points on $\tau^3$-Bench. No research-agent configuration leads across all five tasks.
Research time alone does not account for these outcomes. Across the five tasks, Qwen-3.8-Max uses 81.8 hours, 8.8 hours more than Kimi-K3, but scores lower on four tasks. The following analyses examine validation trajectories, resource use, and component choices behind the final scores.

% --------------------------------------------------

% \vspace{-2mm}
\subsection{Validation Dynamics and Generalization}
\label{sec:validation-trajectories}

\begin{figure}[htbp]
\centering

\includegraphics[width=\linewidth]{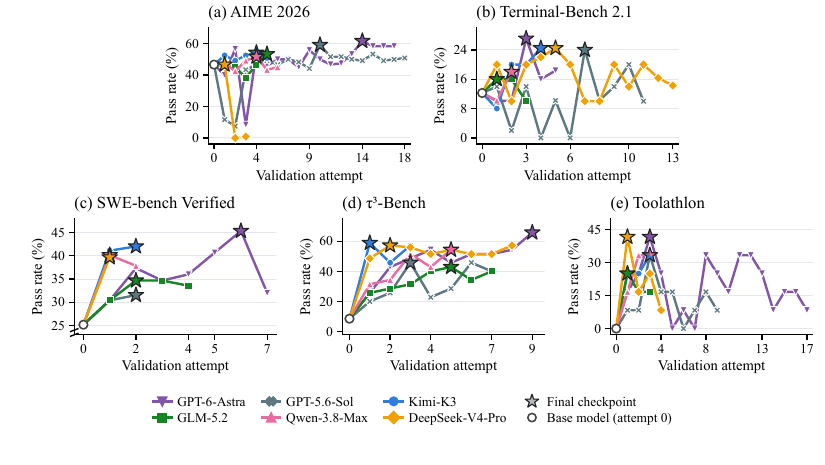}
\vspace{-2.2\baselineskip}
\caption{Each point represents a completed validation evaluation, ordered chronologically within each run. Repeated evaluations of the same candidate are shown separately.}
\label{fig:validation}
\end{figure}

% Figure~\ref{fig:validation} shows that GPT-5.6-Sol reaches the highest AIME~2025 validation score and ties for the highest T-Bench-Lite score after multiple attempts. On the corresponding hidden tests, it scores 50.8\% on AIME~2026, below GLM-5.2's 63.3\%, and 8.9\% on Terminal-Bench~2.1, matching the initial system. GPT-6-Astra leads validation on both SWE-bench Verified and $\tau^3$-Bench. It also leads the SWE hidden test at 53.7\%, whereas Kimi-K3 scores higher on the $\tau^3$-Bench hidden test (52.1\% versus 49.7\%). Validation progress is not monotonic: later DeepSeek-V4-Pro attempts on AIME fall to zero, and GPT-6-Astra's SWE score declines after its peak. On Toolathlon, GPT-6-Astra reaches its best validation score early, with no improvement from its subsequent evaluations. In the runs with later regressions, the selected checkpoint retains an earlier high-scoring candidate. These patterns distinguish finding and preserving a strong validation candidate from producing a system whose gains carry over to hidden tasks.

As shown in Figure~\ref{fig:validation}, the best validation performance does not consistently translate into the best test performance. GPT-5.6-Sol achieves an AIME validation score of 59.2\%, yet scores 50.8\% on the AIME~2026 test set, below GLM-5.2's 63.3\%. GPT-6-Astra leads validation on Terminal-Bench~2.1, SWE-bench Verified, and $\tau^3$-Bench, and ties for the highest Toolathlon validation score. However, it does not consistently lead on the corresponding test sets. On Terminal-Bench~2.1, it scores 13.5\%, compared with Kimi-K3's leading score of 22.4\%. On $\tau^3$-Bench, Kimi-K3 also achieves the highest test score at 52.1\%, ahead of GPT-6-Astra's 49.7\%.  These results show that validation leadership does not reliably predict test leadership.
The validation search is also non-monotonic: later attempts do not necessarily improve on earlier candidates. DeepSeek-V4-Pro's later AIME validation scores fall to zero, while GPT-6-Astra's SWE-bench Verified validation score declines after its peak and its later Toolathlon attempts fail to improve on an early candidate.

% \newpage
\vspace{-2mm}
\noindent\subsection{Token usage analysis}
\vspace{-1mm}
Figure~\ref{fig:output-token-efficiency} compares each research agent's mean generated tokens per run with its macro-averaged test score across the five benchmarks. The target systems produced by each of the six research agents achieve an average score above the initial systems’ 22.7\% across the five benchmarks. But higher token usage does not correspond to higher performance. GLM-5.2 uses the fewest tokens (91.1K per run), while Kimi-K3 achieves the highest score (38.3\%) with 173.2K tokens. These two agents form the observed Pareto frontier. Qwen-3.8-Max, DeepSeek-V4-Pro, and GPT-5.6-Sol use more tokens than Kimi-K3 while scoring lower. GPT-6-Astra achieves the second-highest average score at 37.1\%, but uses by far the most generated tokens. It therefore does not change the observed Pareto frontier formed by GLM-5.2 and Kimi-K3.

% \newpage
% \vspace{-2mm}
\subsection{Component Ablations}
\label{sec:component-ablations}

% \begin{table}[t]
% \caption{Kimi-K3 test scores (\%) under different editable components. D/T denotes data and training; H denotes the inference harness.}
% \label{tab:component-ablation}
% \centering
% \small
% \vspace{6pt}
% \setlength{\tabcolsep}{5pt}
% \begin{tabular}{lcc}
% \toprule
% \textbf{Editable components}
% & \makecell{\textbf{SWE-bench}\\\textbf{Verified}}
% & \makecell{\textbf{Terminal-Bench}\\\textbf{2.1}} \\
% \midrule
% None (baseline) & --   & 8.9  \\
% H only          & 38.0 & 8.0  \\
% D/T only        & 33.0 & 13.0 \\
% D/T/H           & 51.7 & 22.4 \\
% \bottomrule
% \end{tabular}
% \end{table}
% \vspace{-10pt}
\begin{wraptable}{r}{0.5\textwidth}
  \vspace{-2.2\baselineskip}
\centering
\small
% \caption{Kimi-K3 test scores (\%).}
\caption{Kimi-K3 test scores (\%) on SWE-bench Verified and Terminal-Bench~2.1. D/T/H denotes data, trainer, and harness; Kimi-K3 did not edit the harness in the SWE D/T/H run.}
\vspace{6pt}
\label{tab:component-ablation}
\setlength{\tabcolsep}{7pt}
\begin{tabular}{lcc}
\toprule
\textbf{Editable} & \textbf{SWE} & \textbf{Terminal} \\
\midrule
Initial system & 32.9   & 8.9  \\
H only          & 38.0 & 8.9  \\
D/T only        & -- & 13.5 \\
D/T/H           & 51.7 & 22.4 \\
\bottomrule
\end{tabular}
\end{wraptable}
Table~\ref{tab:component-ablation} shows that harness-only editing improves the initial target agent system on SWE-bench Verified but not on Terminal-Bench~2.1. On SWE-bench Verified, the harness-only run raises the test score from 32.9\% to 38.0\%. Kimi-K3's unrestricted run reaches 51.7\% without editing the harness, leaving this harness-based direction unexplored.  On Terminal-Bench~2.1, the harness-only run scores 8.9\%, matching the initial agent system. The data-and-trainer-only run reaches 13.5\%, while the unrestricted run also edits the harness and reaches 22.4\%. 
\FloatBarrier

% Replace unused conference-template example sections with the research analysis.
% \vspace{-1mm}
\section{Agent Behavior Analysis}
\label{sec:analysis}
% We analyze 30 researcher--task runs covering all six researchers and all five tasks. Annotations distinguish \textbf{Data (D)}, \textbf{Trainer (T)}, and \textbf{Harness (H)}. The cohort contains 188 confirmed intervention designs.

\begingroup
\raggedbottom
\makeatletter
\newenvironment{swecase}{\par\noindent\begin{minipage}{\linewidth}\def\@captype{figure}}{\end{minipage}\par\medskip}
\makeatother

\subsection{Research Effort Allocation Across Components}
\label{subsec:effort-allocation}

We define a design as a distinct modification to the training data, trainer, or inference harness that the research agent implements during a run. Repeated submissions of an unchanged candidate count as one design. A design that modifies multiple components contributes to the count for each affected component. Figure~\ref{fig:components} reports the number of designs involving each component. Every research agent modifies data in each task, but the emphasis varies. On AIME~2026, GPT-6-Astra, GPT-5.6-Sol, and GLM-5.2 propose more trainer than data designs, whereas on SWE-bench Verified every research agent except GPT-6-Astra proposes more data designs. Preferences also differ across research agents: Qwen-3.8-Max favors data in all five tasks, while GPT-6-Astra favors the trainer in four of five. GPT-5.6-Sol shifts from trainer-focused designs on AIME~2026 and Terminal-Bench~2.1 to data-focused designs on SWE-bench Verified. Some designs modify multiple components: all seven of GPT-5.6-Sol’s $\tau^3$-Bench designs modify both training data and the trainer. Harness exploration varies within the two tasks that permit it. On SWE-bench Verified, GPT-6-Astra proposes four harness designs, GPT-5.6-Sol proposes two, and Kimi-K3 proposes none. These patterns show that research agents use the editable space differently, both in which components they prioritize and in whether they change components together.
\begin{figure}[htbp]
\centering
% \vspace{-1.2\baselineskip}
\includegraphics[width=\linewidth]{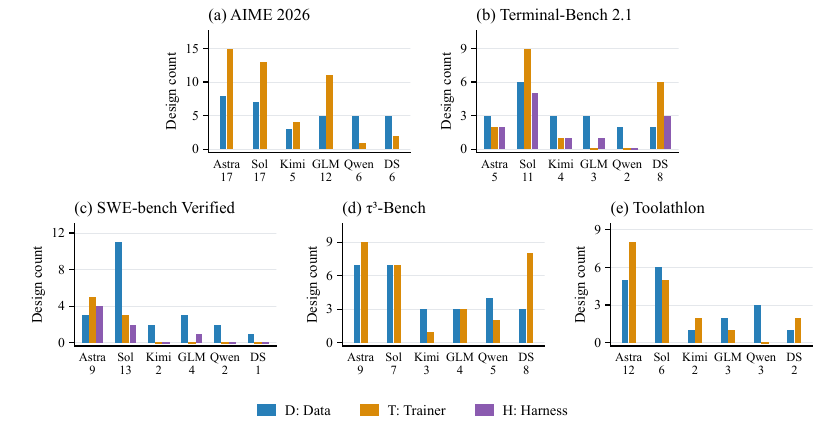}
\vspace{-1.2\baselineskip}
\caption{Number of designs involving training data, the trainer, and the inference harness. A design may involve multiple components. Numbers below the research-agent labels indicate the total number of distinct designs in each run.}
\label{fig:components}
\end{figure}
\subsection{Component Pattern Analysis}

\noindent\textbf{Data strategies.}
Two patterns recur in how research agents construct training data.
\textbf{(1) Trajectory filtering.}
Research agents use similar filtering procedures: retaining successful trajectories, imposing length limits, deduplicating prompts, and removing repetitive execution loops. On SWE-bench Verified, Qwen-3.8-Max and GPT-5.6-Sol remove trajectories with excessive repeated commands. On Toolathlon, Kimi-K3 filters trajectories with repeated tool calls.
\textbf{(2) Format alignment.}
Research agents adapt training examples to the inference harness and evaluation settings. On SWE-bench Verified, Kimi-K3 reformats training prompts to match the inference harness. Qwen-3.8-Max revises tool-call formatting on $\tau^3$-Bench, while GPT-6-Astra selects AIME examples with complete \texttt{\textbackslash boxed\{\}} answers.

\begin{findingbox}
\textbf{Finding 1.} Research agents use similar data preprocessing methods and frequently adjust training-data formats to match task-specific evaluation requirements.
\end{findingbox}

% \noindent\textbf{Trainer strategies.}
% All six research agents use supervised fine-tuning (SFT), with changes at three levels.
% \textbf{(1) Training configuration.} Qwen-3.8-Max and GLM-5.2 test additional epochs on $\tau^3$-Bench, but both candidates score below their preceding validation checkpoints.
% \textbf{(2) Learning signal.} On AIME, GPT-5.6-Sol increases the loss weight on \texttt{</think>}, raising validation performance from 60/120 to 71/120 with the same training examples; GPT-6-Astra makes additional valid assistant turns trainable on $\tau^3$-Bench.
% \textbf{(3) Alternative objectives.} GPT-6-Astra tests GRPO on AIME, but both GRPO branches score 53/120, below the teacher-SFT checkpoint it selects at 60/120.

% \begin{findingbox}
% \textbf{Finding.} Trainer search remains centered on SFT. Task-specific changes to its supervision can help, while additional epochs or a GRPO stage do not reliably improve validation performance in the observed runs.
% \end{findingbox}

\noindent\textbf{Trainer strategies.}
All six research agents use supervised fine-tuning (SFT). Their trainer modifications focus on hyperparameters, loss calculation, and model merging.
\textbf{(1) Training settings.}
Most modifications adjust training hyperparameters. GLM-5.2 tests additional epochs on $\tau^3$-Bench, but validation performance declines. GPT-5.6-Sol tests multiple learning rates on Terminal-Bench, while GPT-6-Astra reduces the training batch size on SWE-bench Verified. Qwen-3.8-Max reduces the pipeline parallel size on $\tau^3$-Bench to accelerate training.
\textbf{(2) Loss calculation.}
On AIME, GPT-5.6-Sol increases the loss weight on \texttt{</think>} to encourage shorter outputs, improving the validation score from 60/120 to 71/120 with the same examples. GPT-6-Astra adds supervision on valid historical assistant turns in $\tau^3$-Bench examples to obtain more training targets.
\textbf{(3) Model merging.}
On AIME, Kimi-K3 compares two-model and three-model merges and selects the two-model merge after it scores higher on validation. DeepSeek-V4-Pro tests different interpolation coefficients on Terminal-Bench to select the final model weights. GPT-6-Astra merges two trained models on $\tau^3$-Bench, achieving the highest validation score in its run.

\begin{figure}[t]
\centering
\begin{tikzpicture}[
  stage/.style={
    draw=FigureBlue!45,
    fill=FigureBlue!3,
    rounded corners=2pt,
    text width=0.82\linewidth,
    align=left,
    inner sep=6pt,
    font=\small
  },
  omission/.style={
    text width=0.82\linewidth,
    align=center,
    inner sep=1pt,
    font=\footnotesize,
    text=FigureBlue
  },
  marker/.style={
    circle,
    fill=FigureBlue,
    inner sep=0pt,
    minimum size=5pt
  }
]

\node[stage, anchor=north west] (s1) at (0.5cm,0) {
  \textbf{Decision 1: Train on successful repair demonstrations}\\
To improve repository repair, select demonstrations for 1,536 resolved issues from the Orchard training dataset  and train for one epoch.
  Keep the original harness.
  Validation: \textbf{45 solved tasks}.
};

\node[omission, anchor=north west] (o1)
  at ([yshift=-1mm]s1.south west) {
  $\vdots$\quad 
};

\node[stage, anchor=north west] (s3)
  at ([yshift=-1mm]o1.south west) {
  \textbf{Decision 3: Fix Python environment selection in the harness}\\
  Validation traces from the first trained model reveal
  a Python interpreter mismatch.
  Revise the harness to use the repository's existing Python environment
  while preserving the agent's own interpreter.
  With model weights fixed, the number of validation tasks solved rises from \textbf{45 to 55}.
  Full validation: \textbf{55/150}.
};

\node[omission, anchor=north west] (o2)
  at ([yshift=-1mm]s3.south west) {
  $\vdots$\quad 
};

\node[stage, anchor=north west] (s7)
  at ([yshift=-1mm]o2.south west) {
  \textbf{Decision 7: Test broader training-data coverage}\\
  Test whether expanding Orchard data from 1,536 to 4,096 issues
  improves task success.
  Evaluate the model trained for one epoch on the expanded data
  with the revised harness.
  Validation: \textbf{54/150}.
};

\node[omission, anchor=north west] (o3)
  at ([yshift=-1mm]s7.south west) {
  $\vdots$\quad 
};

\node[stage, fill=FigureBlue!8, anchor=north west] (s9)
  at ([yshift=-1mm]o3.south west) {
  \textbf{Decision 9: Test additional training on the same data}\\
  Test whether a second training epoch improves task success.
  Keep the same 4,096 Orchard issues and revised harness;
  evaluate the model trained for two epochs.
  Validation: \textbf{68/150}.
  Select this system; test score: \textbf{53.7\%}.
};

\coordinate (m1) at ([xshift=-4mm]s1.west);
\coordinate (m3) at ([xshift=-4mm]s3.west);
\coordinate (m7) at ([xshift=-4mm]s7.west);
\coordinate (m9) at ([xshift=-4mm]s9.west);

\draw[FigureBlue, densely dotted, line width=0.7pt]
  (m1) -- (m3) -- (m7) -- (m9);

\foreach \m in {m1,m3,m7,m9}
  \node[marker] at (\m) {};

\end{tikzpicture}
\caption{Case study of GPT-6-Astra's research trajectory on SWE-bench Verified. Selected decisions illustrate how validation feedback guides changes to training data, the trainer, and the inference harness. Ellipses indicate omitted decisions.}
\label{fig:swe-case-astra}
\end{figure}
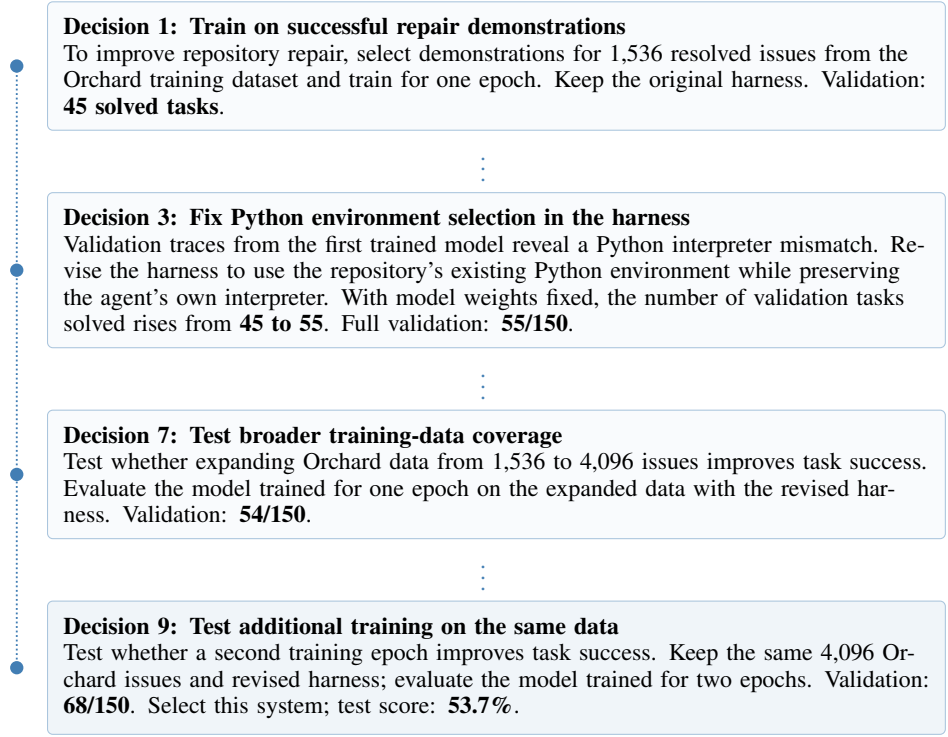

\begin{findingbox}
\textbf{Finding 2.} 
Trainer modifications concentrate on hyperparameter tuning. Changes to loss calculation are less frequent. Several agents also use model merging to combine multiple candidate models. 
\end{findingbox}

\noindent\textbf{Harness strategies.}
On SWE-bench Verified and Terminal-Bench~2.1, harness modifications focus on execution control and parsing recovery.
\textbf{(1) Execution control.} GPT-5.6-Sol adds guards against repeated commands and unchanged patches on SWE-bench Verified, reducing median API calls from 163 to 96.5 in a paired 30-task comparison. On Terminal-Bench~2.1, Kimi-K3 adds a completion checklist that prompts the target agent to verify task requirements before declaring completion.
\textbf{(2) Output recovery.} Research agents modify output parsers to recover from malformed responses. On Terminal-Bench~2.1, DeepSeek-V4-Pro adds automatic repairs for escaping issues, missing colons, truncated outputs, and trailing commas. GPT-6-Astra also adds automatic repair of invalid escape sequences and revises JSON extraction logic to better identify structured responses.

\begin{findingbox}
\textbf{Finding 3.} Research agents primarily make incremental modifications to execution control and output recovery, leaving the overall inference harness structure unchanged.
\end{findingbox}

\subsection{Case Study}
\label{sec:research-actions}
% \begin{swecase}
% \label{app:case-swe-astra}
% The GPT-6-Astra run on SWE-bench Verified illustrates a feedback-guided shift from training to harness revision and back (Figure~\ref{fig:swe-case-astra}). After initial training, execution traces reveal a Python interpreter mismatch between tool commands and repository dependencies. The research agent revises the harness to use the repository's existing environment while preserving the agent's own interpreter. With model weights fixed, this revision raises solves from 45 to 55 on validation tasks. The agent then retains the
% revised harness while changing the training data and trainer; the selected system solves 68/150 validation tasks and scores 53.7\% on the hidden test.

% \par\smallskip
% \centering
% \includegraphics[width=\linewidth]{figures/swe_case_astra.pdf}
% \caption{GPT-6-Astra's SWE-bench Verified research trajectory. The research agent tests an environment-selecting inference harness with model weights fixed.}
% \label{fig:swe-case-astra}
% \end{swecase}

The GPT-6-Astra run on SWE-bench Verified illustrates how validation
feedback guides changes to training data, the trainer, and the
inference harness (Figure~\ref{fig:swe-case-astra}).
After the research agent trains the model on demonstrations for 1,536 issues from Orchard, a software-repair training dataset, execution traces reveal a Python interpreter mismatch.
The research agent revises the harness to use the repository's
existing Python environment while preserving the agent's own interpreter. With model weights fixed, this revision increases the number of validation tasks solved from 45 to 55. The agent also expands training to 4,096 issues and evaluates the resulting checkpoints with the revised harness. The step-128 and step-256 checkpoints solve 54 and 68 of the validation tasks, respectively. The agent selects the step-256 model with the revised harness, achieving 53.7\% on the hidden test.
\section{Conclusion}
In this paper, we introduced \textsc{VibeTraining-Bench}, a benchmark designed to evaluate research agents' cross-component co-design capabilities across diverse domains. In each task, a research agent begins with an initial target agent system and iteratively modifies the training data, the trainer, and the task-specific inference harness. Finally, the research agent selects and submits a final agent system for evaluation on test tasks. 
We develop a reliable execution platform to guarantee evaluation integrity. To prevent research agents from reward hacking, our platform implements three levels of safeguards by isolating test data, separating research and grading containers, and restricting external network access. 
Our evaluation results show that frontier research agents can substantially improve target agent systems, in some cases approaching the performance of the system with human post-training. 
%However, an in-depth analysis of research trajectories and component ablations reveals critical limitations
However, an in-depth analysis of research agents' trajectories and component ablation experiments reveals that these research agents still exhibit limitations. 
%They explore improvable system components unevenly, and validation gains do not consistently translate into improvements on test tasks. 
%There is still room for improvement in their Data–Trainer–Harness co-design capability and iterative refinement toward achieving cumulative performance gains. 
There is still room to improve their ability to co-design and iteratively refine data, trainer, and harness to achieve cumulative performance gains. 
%These findings delineate both the capabilities and bottlenecks of current research agents. 
We hope that \textsc{VibeTraining-Bench} will serve as a unified platform for the community to develop more reliable and effective research agents for recursive self-improvement. 
\label{main-text-end}
\clearpage
\section*{AI Use Statement}
In this work, we used generative AI tools to draft and polish portions of the manuscript, and to improve clarity, grammar, and readability. We did not use generative AI tools for the research tasks requiring disclosure under the ICLR policy. All AI-assisted text was reviewed and edited by the authors, who take responsibility for the final content of this work, including all claims and results.

% \section*{Ethics Statement}
% The benchmark studies agents that can modify and execute training and inference software. Its intended use is controlled capability evaluation in isolated environments with explicit data and evaluation boundaries. Improved automation can also increase the scale at which unreliable code or misleading experiments are produced. We therefore report unsuccessful interventions and measurement limitations alongside favorable results. Dataset licensing, resource provenance, and the release contents for the remaining task families require verification when their manifests become available.

\bibliography{iclr2027_conference}

@inproceedings{huang2026r,
  title={R-zero: Self-evolving reasoning llm from zero data},
  author={Huang, Chengsong and Yu, Wenhao and Wang, Xiaoyang and Zhang, Hongming and Li, Zongxia and Li, Ruosen and Huang, Jiaxin and Mi, Haitao and Yu, Dong},
  booktitle={International Conference on Learning Representations},
  volume={2026},
  pages={130770--130790},
  year={2026}
}

@article{zhao2026absolute,
  title={Absolute zero: Reinforced self-play reasoning with zero data},
  author={Zhao, Andrew and Wu, Yiran and Wu, Tong and Xu, Quentin and Yue, Yang and Lin, Matthieu and Wang, Shenzhi and Wu, Qingyun and Zheng, Zilong and Huang, Gao},
  journal={Advances in Neural Information Processing Systems},
  volume={38},
  pages={105816--105879},
  year={2026}
}

@inproceedings{yin2025godel,
  title={G{\"o}del agent: A self-referential agent framework for recursively self-improvement},
  author={Yin, Xunjian and Wang, Xinyi and Pan, Liangming and Lin, Li and Wan, Xiaojun and Wang, William Yang},
  booktitle={Proceedings of the 63rd Annual Meeting of the Association for Computational Linguistics (Volume 1: Long Papers)},
  pages={27890--27913},
  year={2025}
}

@article{fang2026llmzero,
  title={LLMZero: Discovering Adaptive Training Strategies for RL Post-Training via LLM Agents},
  author={Fang, Haoyang and Zhu, Wei and Han, Boran and Zhang, Alex and Pan, Zhenyu and Yang, Shuo and Zhang, Shuai and Gai, Jiading and Tang, Peng and Hu, Cuixiong and others},
  journal={arXiv preprint arXiv:2606.18388},
  year={2026}
}

@article{lu2024discovering,
  title={Discovering preference optimization algorithms with and for large language models},
  author={Lu, Chris and Holt, Samuel and Fanconi, Claudio and Chan, Alex J and Foerster, Jakob and van der Schaar, Mihaela and Lange, Robert T},
  journal={Advances in Neural Information Processing Systems},
  volume={37},
  pages={86528--86573},
  year={2024}
}

@inproceedings{zhang2026darwin,
  title={Darwin G{\"o}del machine: open-ended evolution of self-improving agents},
  author={Zhang, Jenny and Hu, Shengran and Lu, Cong and Lange, Robert and Clune, Jeff},
  booktitle={International Conference on Learning Representations},
  volume={2026},
  pages={104223--104294},
  year={2026}
}

@article{chen2026harnessx,
  title={Harnessx: A composable, adaptive, and evolvable agent harness foundry},
  author={Chen, Tingyang and Lu, Shuo and Zhao, Kang and Meng, Weicheng and Teng, Hanlin and Li, Tianhao and Li, Chao and Liu, Xule and Liang, Jian and Zhang, Zhizhong and others},
  journal={arXiv preprint arXiv:2606.14249},
  year={2026}
}

@article{chen2026evotrainer,
  title={EvoTrainer: Co-Evolving LLM Policies and Training Harnesses for Autonomous Agentic Reinforcement Learning},
  author={Chen, Guhong and Shi, Yingcheng and Li, Yongbin and Li, Binhua and Xu, Xander and Wei, Hu and Ni, Shiwen and Yang, Min and Ye, Jieping},
  journal={arXiv preprint arXiv:2606.03108},
  year={2026}
}

@article{huang2023mlagentbench,
  title={Mlagentbench: Evaluating language agents on machine learning experimentation},
  author={Huang, Qian and Vora, Jian and Liang, Percy and Leskovec, Jure},
  journal={arXiv preprint arXiv:2310.03302},
  year={2023}
}

@article{wijk2024re,
  title={Re-bench: Evaluating frontier ai r\&d capabilities of language model agents against human experts},
  author={Wijk, Hjalmar and Lin, Tao and Becker, Joel and Jawhar, Sami and Parikh, Neev and Broadley, Thomas and Chan, Lawrence and Chen, Michael and Clymer, Josh and Dhyani, Jai and others},
  journal={arXiv preprint arXiv:2411.15114},
  year={2024}
}

@inproceedings{chan2025mle,
  title={Mle-bench: Evaluating machine learning agents on machine learning engineering},
  author={Chan, Jun Shern and Chowdhury, Neil and Jaffe, Oliver and Aung, James and Sherburn, Dane and Mays, Evan and Starace, Giulio and Liu, Kevin and Maksin, Leon and Patwardhan, Tejal and others},
  booktitle={International Conference on Learning Representations},
  volume={2025},
  pages={50466--50494},
  year={2025}
}

@article{meng2026rsibench,
  title={RSIBench-Data: Benchmarking data-centric research for recursive self-improvement},
  author={Meng, Fanqing and Du, Lingxiao and Chen, Qiguang and Zhao, Ziqi and Lu, Haocheng and Hu, Mengkang and Shieh, Michael Qizhe},
  journal={arXiv preprint arXiv:2607.25886},
  year={2026}
}

@article{lyu2026mls,
  title={MLS-Bench: A holistic and rigorous assessment of AI systems on building better AI},
  author={Lyu, Bohan and Yang, Yucheng and Huang, Siqiao and Zhang, Jiaru and Xu, Qixin and Li, Xinghan and Han, Xinyang and Zhang, Yicheng and Zhang, Huaqing and Huang, Runhan and others},
  journal={arXiv preprint arXiv:2605.08678},
  year={2026}
}

@article{chi2026ai4ai,
  title={Ai4ai-bench: Benchmarking llm agents in algorithmic design for recursive self-improvement},
  author={Chi, Yizhe and Li, Wenyi and Hong, Deyao and Wang, Xiaoqiu and Gao, Mingju and Yang, Kaisen and He, Bingxiang and Zheng, Youjie and Xiao, Calvin and Na, Qinhuai},
  journal={arXiv preprint arXiv:2608.20318},
  year={2026}
}

@article{starace2025paperbench,
  title={PaperBench: evaluating AI's ability to replicate AI research},
  author={Starace, Giulio and Jaffe, Oliver and Sherburn, Dane and Aung, James and Chan, Jun Shern and Maksin, Leon and Dias, Rachel and Mays, Evan and Kinsella, Benjamin and Thompson, Wyatt and others},
  journal={arXiv preprint arXiv:2504.01848},
  year={2025}
}

@article{yang2026frontis,
  title={Frontis-MA1: Training an AI4AI model towards recursive self-improvement in machine learning engineering},
  author={Yang, Junlin and Jiang, Che and Fu, Yu and Luo, Tianwei and Ren, Can and Wang, Weizhi and Zhao, Kaikai and Liu, Hongyi and Zuo, Yuxin and Wang, Yuru and others},
  journal={arXiv preprint arXiv:2607.28568},
  year={2026}
}

@article{chen2026co,
  title={Co-harness: Co-evolving harnesses and model weights for llm agents},
  author={Chen, Zhengyu and Xiao, Teng and Zhu, Huaisheng and Yuan, Yige and Zhang, Luan and Wang, Jingang},
  journal={arXiv preprint arXiv:2607.22688},
  year={2026}
}

@article{kang2026can,
  title={Can Generalist Agents Automate Data Curation?},
  author={Kang, Feiyang and Li, Hanze and Nguyen, Adam and Dabas, Mahavir and Ma, Jiaqi W and Sala, Frederic and Song, Dawn and Jia, Ruoxi},
  journal={arXiv preprint arXiv:2606.04261},
  year={2026}
}

@article{zweiger2026self,
  title={Self-adapting language models},
  author={Zweiger, Adam and Pari, Jyo and Guo, Han and Kim, Yoon and Agrawal, Pulkit},
  journal={Advances in Neural Information Processing Systems},
  volume={38},
  pages={74084--74115},
  year={2026}
}

@article{rank2026posttrainbench,
  title={PostTrainBench: Can LLM Agents Automate LLM Post-Training?},
  author={Rank, Ben and Bhatnagar, Hardik and Prabhu, Ameya and Eisenberg, Shira and Nguyen, Karina and Bethge, Matthias and Andriushchenko, Maksym},
  journal={arXiv preprint arXiv:2603.08640},
  year={2026}
}

@article{ursekar2026harnessopt,
  title={HarnessOpt-Bench: Evaluating LLMs at Harness Optimization},
  author={Ursekar, Varun and Shanker, Apaar and Maurya, Yash and Yasser, Shehab and Kalmath, Vijay S and Chatrath, Veronica and Xue, Yuan},
  journal={arXiv preprint arXiv:2608.06301},
  year={2026}
}

@article{202608.0051,
  doi = {10.20944/preprints202608.0051.v1},
  url = {https://www.preprints.org/manuscript/202608.0051},
  year = 2026,
  month = {August},
  publisher = {Preprints},
  author = {Shuaiqi Liu and Zhengkai Lin and Yuxiang Zhang and Yuanyi Ren and Yue Wu and Yongbin Li and Zheng Wang and Zhihang Fu and Jieping Ye},
  title = {The Path to Recursive Self-Improving Agents: Foundation, Framework, and Future Directions},
  journal = {Preprints}
}

@article{chen2026agent,
  title={Agent\^{} 2 RL-Bench: Can LLM Agents Engineer Agentic RL Post-Training?},
  author={Chen, Wanyi and Yang, Xiao and Yang, Xu and Sha, Tianming and Li, Qizheng and Wang, Zhuo and Xian, Bowen and Kong, Fang and Liu, Weiqing and Bian, Jiang},
  journal={arXiv preprint arXiv:2604.10547},
  year={2026}
}

@article{huang2026evo,
  title={Evo-Bench: Can Language Models Improve Agent Harness?},
  author={Huang, Lisheng and Yang, Chen and Zhou, Hao and Song, Huatong and Chen, Zongchao and Le, Ran and Song, Yang and Zhao, Wayne Xin and Zhang, Tao},
  journal={arXiv preprint arXiv:2608.09096},
  year={2026}
}
\bibliographystyle{iclr2027_conference}
\clearpage
\appendix

\providecolor{FigureBlue}{HTML}{427EB4}

\section{Terminal-Bench Case Studies}
\label{app:trajectory}

\subsection{Case Study: Kimi-K3 on Terminal-Bench} 

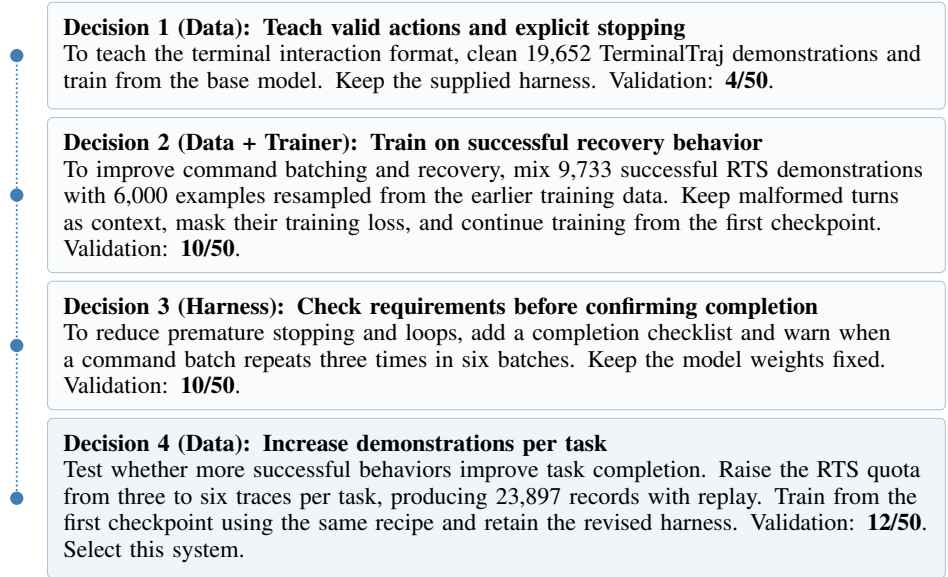
\begin{figure}[htbp]
\centering
\begin{tikzpicture}[
  stage/.style={draw=FigureBlue!45,fill=FigureBlue!3,rounded corners=2pt,
    text width=0.82\linewidth,align=left,inner sep=6pt,font=\small},
  omission/.style={text width=0.82\linewidth,align=center,inner sep=1pt,
    font=\footnotesize,text=FigureBlue},
  marker/.style={circle,fill=FigureBlue,inner sep=0pt,minimum size=5pt}
]
\node[stage,anchor=north west] (d1) at (0.5cm,0) {
  \textbf{Decision 1 (Data): Teach valid actions and explicit stopping}\\
  To teach the terminal interaction format, clean 19,652 TerminalTraj demonstrations and train from the base model. Keep the supplied harness. Validation: \textbf{4/50}.
};

\node[stage,anchor=north west] (d2) at ([yshift=-1mm]d1.south west) {
  \textbf{Decision 2 (Data + Trainer): Train on successful recovery behavior}\\
  To improve command batching and recovery, mix 9,733 successful RTS demonstrations with 6,000 examples resampled from the earlier training data. Keep malformed turns as context, mask their training loss, and continue training from the first checkpoint. Validation: \textbf{10/50}.
};

\node[stage,anchor=north west] (d3) at ([yshift=-1mm]d2.south west) {
  \textbf{Decision 3 (Harness): Check requirements before confirming completion}\\
  To reduce premature stopping and loops, add a completion checklist and warn when a command batch repeats three times in six batches. Keep the model weights fixed. Validation: \textbf{10/50}.
};

\node[stage,fill=FigureBlue!8,anchor=north west] (d4) at ([yshift=-1mm]d3.south west) {
  \textbf{Decision 4 (Data): Increase demonstrations per task}\\
  Test whether more successful behaviors improve task completion. Raise the RTS quota from three to six traces per task, producing 23,897 records with replay. Train from the first checkpoint using the same recipe and retain the revised harness. Validation: \textbf{12/50}. Select this system.
};

\coordinate (m1) at ([xshift=-4mm]d1.west);
\coordinate (m2) at ([xshift=-4mm]d2.west);
\coordinate (m3) at ([xshift=-4mm]d3.west);
\coordinate (m4) at ([xshift=-4mm]d4.west);
\draw[FigureBlue,densely dotted,line width=0.7pt] (m1) -- (m2) -- (m3) -- (m4);
\node[marker] at (m1) {};
\node[marker] at (m2) {};
\node[marker] at (m3) {};
\node[marker] at (m4) {};
\end{tikzpicture}
\caption{Kimi-K3 on Terminal-Bench. The selected system combines the expanded data mixture with the revised harness.}
\label{fig:terminal-case-kimi}
\end{figure}

Kimi-K3 first trains the model to use the terminal interaction format, then uses validation failures to target recovery and completion behavior (Figure~\ref{fig:terminal-case-kimi}). Recovery-focused continuation raises solves from 4/50 to 10/50. A fixed-weight harness revision leaves that total unchanged, although its completion checklist is followed by resumed commands in 28 of 66 completion events. Kimi-K3 then expands demonstrations per task and selects the resulting model with the revised harness at 12/50.

\subsection{Case Study: Qwen-3.8-Max on Terminal-Bench}
\label{app:case-terminal-Qwen-3.8-Max}

\begin{figure}[htbp]
\centering
\begin{tikzpicture}[
  stage/.style={draw=FigureBlue!45,fill=FigureBlue!3,rounded corners=2pt,
    text width=0.82\linewidth,align=left,inner sep=6pt,font=\small},
  omission/.style={text width=0.82\linewidth,align=center,inner sep=1pt,
    font=\footnotesize,text=FigureBlue},
  marker/.style={circle,fill=FigureBlue,inner sep=0pt,minimum size=5pt}
]
\node[stage,anchor=north west] (d1) at (0.5cm,0) {
  \textbf{Decision 1 (Data): Teach the terminal interface from completed trajectories}\\
  To teach valid actions and stopping, normalize JSON responses and mix 8,000 successful RTS with 8,000 TerminalTraj demonstrations. Train from the base using the supplied recipe and harness. Validation: \textbf{5/50}.
};

\node[stage,fill=FigureBlue!8,anchor=north west] (d2) at ([yshift=-1mm]d1.south west) {
  \textbf{Decision 2 (Data): Emphasize explicit verification in the curriculum}\\
  Test whether verification demonstrations reduce premature completion. Mix 5,500 verification RTS, 6,000 other RTS, and 4,500 TerminalTraj examples; cap TerminalTraj at 40 assistant turns. Keep 16,000 records, 250 training steps, and the harness. Validation: \textbf{9/50}. Select this system.
};

\coordinate (m1) at ([xshift=-4mm]d1.west);
\coordinate (m2) at ([xshift=-4mm]d2.west);
\draw[FigureBlue,densely dotted,line width=0.7pt] (m1) -- (m2);
\node[marker] at (m1) {};
\node[marker] at (m2) {};
\end{tikzpicture}
\caption{Qwen-3.8-Max on Terminal-Bench. The data curriculum changes while the training recipe and harness remain fixed.}
\label{fig:terminal-case-qwen}
\end{figure}
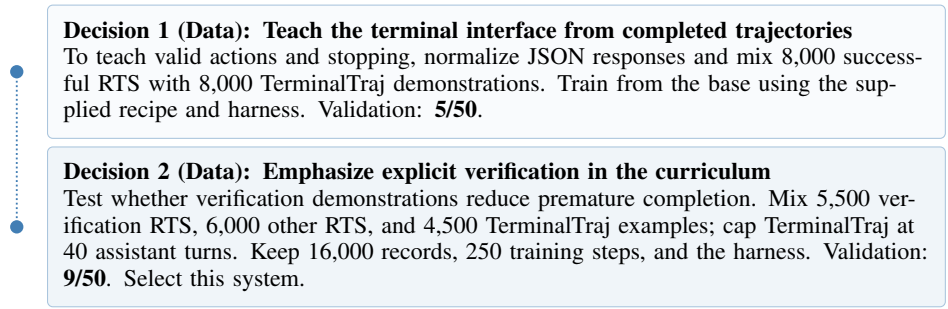

Qwen-3.8-Max responds to premature completion and long loops by changing the demonstrated behavior (Figure~\ref{fig:terminal-case-qwen}). It replaces a broad keyword heuristic with a filter for explicit tests, output checks, and parsing commands, then gives these examples more curriculum weight and limits long TerminalTraj records. Both models train afresh from the same base on 16,000 examples with the same 250-step recipe and harness. The revised curriculum is selected after validation changes from 5/50 to 9/50.

\subsection{Case Study: GPT-5.6-Sol on Terminal-Bench}
\label{app:case-terminal-sol}
GPT-5.6-Sol\textquotesingle{}s validation failures expose a mismatch between intended and executed training (Figure~\ref{fig:terminal-case-sol}). In Decisions 2, 4, and 6, the agent intends to continue training from a trained checkpoint but instead initializes the model randomly: the loader receives an iteration directory rather than its parent checkpoint root. Correcting the path preserves the data, learning rate, and harness of the preceding comparison and raises validation from 0/50 to 12/50. Later checkpoint, data, and training branches do not exceed this result; GPT-5.6-Sol retains the correctly initialized continuation.
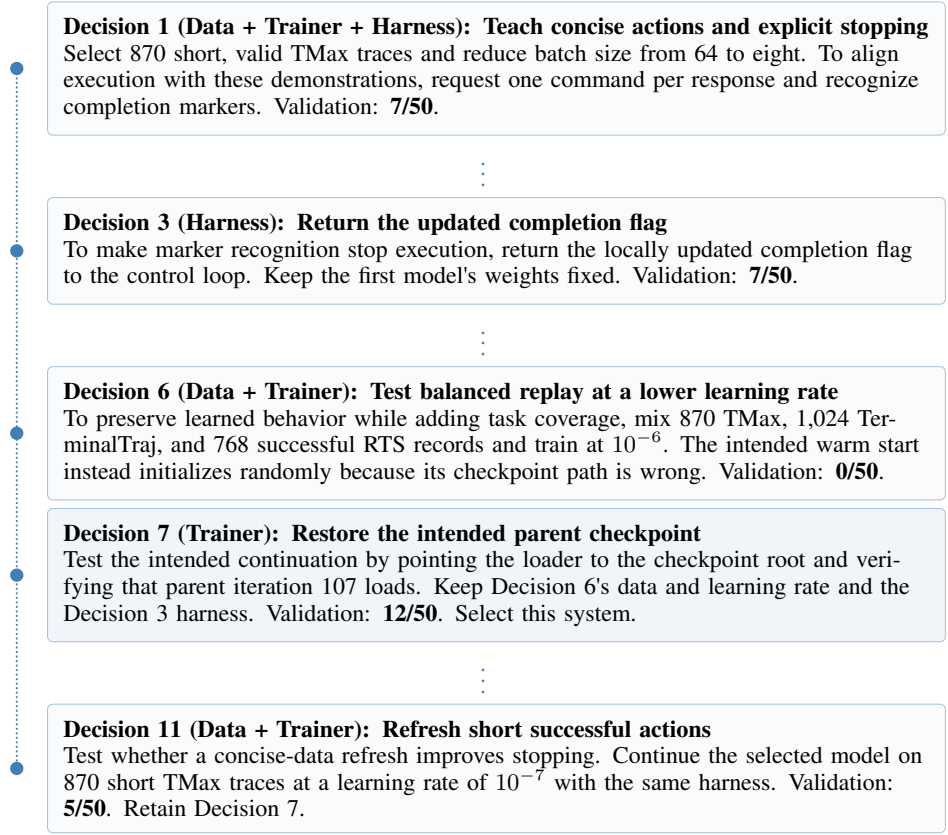
\begin{figure}[htbp]
\centering
\begin{tikzpicture}[
  stage/.style={draw=FigureBlue!45,fill=FigureBlue!3,rounded corners=2pt,
    text width=0.82\linewidth,align=left,inner sep=6pt,font=\small},
  omission/.style={text width=0.82\linewidth,align=center,inner sep=1pt,
    font=\footnotesize,text=FigureBlue},
  marker/.style={circle,fill=FigureBlue,inner sep=0pt,minimum size=5pt}
]
\node[stage,anchor=north west] (d1) at (0.5cm,0) {
  \textbf{Decision 1 (Data + Trainer + Harness): Teach concise actions and explicit stopping}\\
  Select 870 short, valid TMax traces and reduce batch size from 64 to eight. To align execution with these demonstrations, request one command per response and recognize completion markers. Validation: \textbf{7/50}.
};

\node[omission,anchor=north west] (o3) at ([yshift=-1mm]d1.south west) {$\vdots$};
\node[stage,anchor=north west] (d3) at ([yshift=-1mm]o3.south west) {
  \textbf{Decision 3 (Harness): Return the updated completion flag}\\
  To make marker recognition stop execution, return the locally updated completion flag to the control loop. Keep the first model\textquotesingle{}s weights fixed. Validation: \textbf{7/50}.
};

\node[omission,anchor=north west] (o6) at ([yshift=-1mm]d3.south west) {$\vdots$};
\node[stage,anchor=north west] (d6) at ([yshift=-1mm]o6.south west) {
  \textbf{Decision 6 (Data + Trainer): Test balanced replay at a lower learning rate}\\
  To preserve learned behavior while adding task coverage, mix 870 TMax, 1,024 TerminalTraj, and 768 successful RTS records and train at $10^{-6}$. The intended warm start instead initializes randomly because its checkpoint path is wrong. Validation: \textbf{0/50}.
};

\node[stage,fill=FigureBlue!8,anchor=north west] (d7) at ([yshift=-1mm]d6.south west) {
  \textbf{Decision 7 (Trainer): Restore the intended parent checkpoint}\\
  Test the intended continuation by pointing the loader to the checkpoint root and verifying that parent iteration 107 loads. Keep Decision 6\textquotesingle{}s data and learning rate and the Decision 3 harness. Validation: \textbf{12/50}. Select this system.
};

\node[omission,anchor=north west] (o10) at ([yshift=-1mm]d7.south west) {$\vdots$};
% \node[stage,anchor=north west] (d10) at ([yshift=-1mm]o10.south west) {
%   \textbf{Decision 10 (Trainer): Test another epoch on the same data}\\
%   Test whether another pass consolidates terminal behavior. Continue the selected model on the same 2,662 records at a learning rate of $2\times10^{-7}$, retaining the harness. Validation: \textbf{10/50}.
% };

\node[stage,anchor=north west] (d11) at ([yshift=-1mm]o10.south west) {
  \textbf{Decision 11 (Data + Trainer): Refresh short successful actions}\\
  Test whether a concise-data refresh improves stopping. Continue the selected model on 870 short TMax traces at a learning rate of $10^{-7}$ with the same harness. Validation: \textbf{5/50}. Retain Decision 7.
};

\coordinate (m1) at ([xshift=-4mm]d1.west);
\coordinate (m3) at ([xshift=-4mm]d3.west);
\coordinate (m6) at ([xshift=-4mm]d6.west);
\coordinate (m7) at ([xshift=-4mm]d7.west);
% \coordinate (m10) at ([xshift=-4mm]d10.west);
\coordinate (m11) at ([xshift=-4mm]d11.west);
\draw[FigureBlue,densely dotted,line width=0.7pt] (m1) -- (m3) -- (m6) -- (m7)  -- (m11);
\node[marker] at (m1) {};
\node[marker] at (m3) {};
\node[marker] at (m6) {};
\node[marker] at (m7) {};
% \node[marker] at (m10) {};
\node[marker] at (m11) {};
\end{tikzpicture}
\caption{GPT-5.6-Sol on Terminal-Bench. Decisions 6 and 7 hold data, learning rate, and harness fixed while correcting initialization.}
\label{fig:terminal-case-sol}
\end{figure}

\subsection{Case Study: GLM-5.2 on Terminal-Bench}
\label{app:case-terminal-glm}

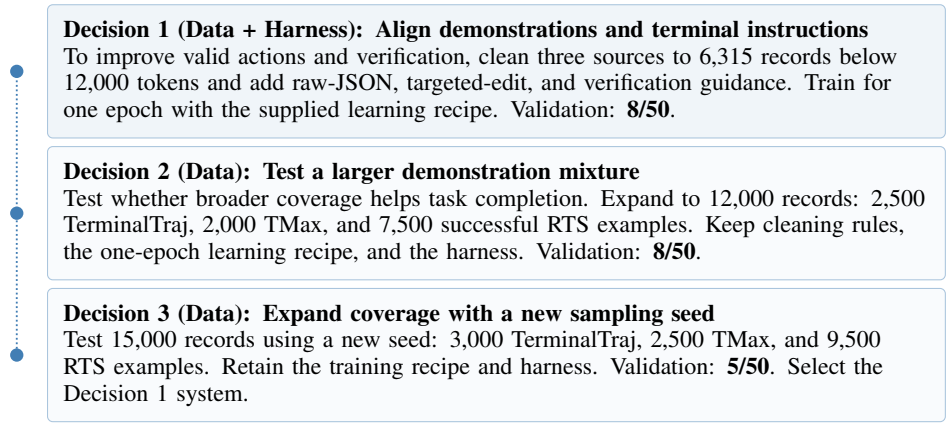
\begin{figure}[htbp]
\centering
\begin{tikzpicture}[
  stage/.style={draw=FigureBlue!45,fill=FigureBlue!3,rounded corners=2pt,
    text width=0.82\linewidth,align=left,inner sep=6pt,font=\small},
  omission/.style={text width=0.82\linewidth,align=center,inner sep=1pt,
    font=\footnotesize,text=FigureBlue},
  marker/.style={circle,fill=FigureBlue,inner sep=0pt,minimum size=5pt}
]
\node[stage,fill=FigureBlue!8,anchor=north west] (d1) at (0.5cm,0) {
  \textbf{Decision 1 (Data + Harness): Align demonstrations and terminal instructions}\\
  To improve valid actions and verification, clean three sources to 6,315 records below 12,000 tokens and add raw-JSON, targeted-edit, and verification guidance. Train for one epoch with the supplied learning recipe. Validation: \textbf{8/50}.
};

\node[stage,anchor=north west] (d2) at ([yshift=-1mm]d1.south west) {
  \textbf{Decision 2 (Data): Test a larger demonstration mixture}\\
  Test whether broader coverage helps task completion. Expand to 12,000 records: 2,500 TerminalTraj, 2,000 TMax, and 7,500 successful RTS examples. Keep cleaning rules, the one-epoch learning recipe, and the harness. Validation: \textbf{8/50}.
};

\node[stage,anchor=north west] (d3) at ([yshift=-1mm]d2.south west) {
  \textbf{Decision 3 (Data): Expand coverage with a new sampling seed}\\
  Test 15,000 records using a new seed: 3,000 TerminalTraj, 2,500 TMax, and 9,500 RTS examples. Retain the training recipe and harness. Validation: \textbf{5/50}. Select the Decision 1 system.
};

\coordinate (m1) at ([xshift=-4mm]d1.west);
\coordinate (m2) at ([xshift=-4mm]d2.west);
\coordinate (m3) at ([xshift=-4mm]d3.west);
\draw[FigureBlue,densely dotted,line width=0.7pt] (m1) -- (m2) -- (m3);
\node[marker] at (m1) {};
\node[marker] at (m2) {};
\node[marker] at (m3) {};
\end{tikzpicture}
\caption{GLM-5.2 on Terminal-Bench. Data expansion fails to improve on the first selected model--harness pair.}
\label{fig:terminal-case-glm}
\end{figure}

GLM-5.2 tests data expansion after aligning demonstrations and terminal instructions (Figure~\ref{fig:terminal-case-glm}). Increasing the mixture from 6,315 to 12,000 records leaves validation at 8/50; expanding to 15,000 records with a new seed lowers it to 5/50. The one-epoch recipe and harness are retained throughout these expansions. GLM-5.2 selects the first model. The larger step counts follow the larger datasets rather than an independent change to the learning recipe.

\subsection{Case Study: DeepSeek-V4-Pro on Terminal-Bench}

\label{app:case-terminal-deepseek}

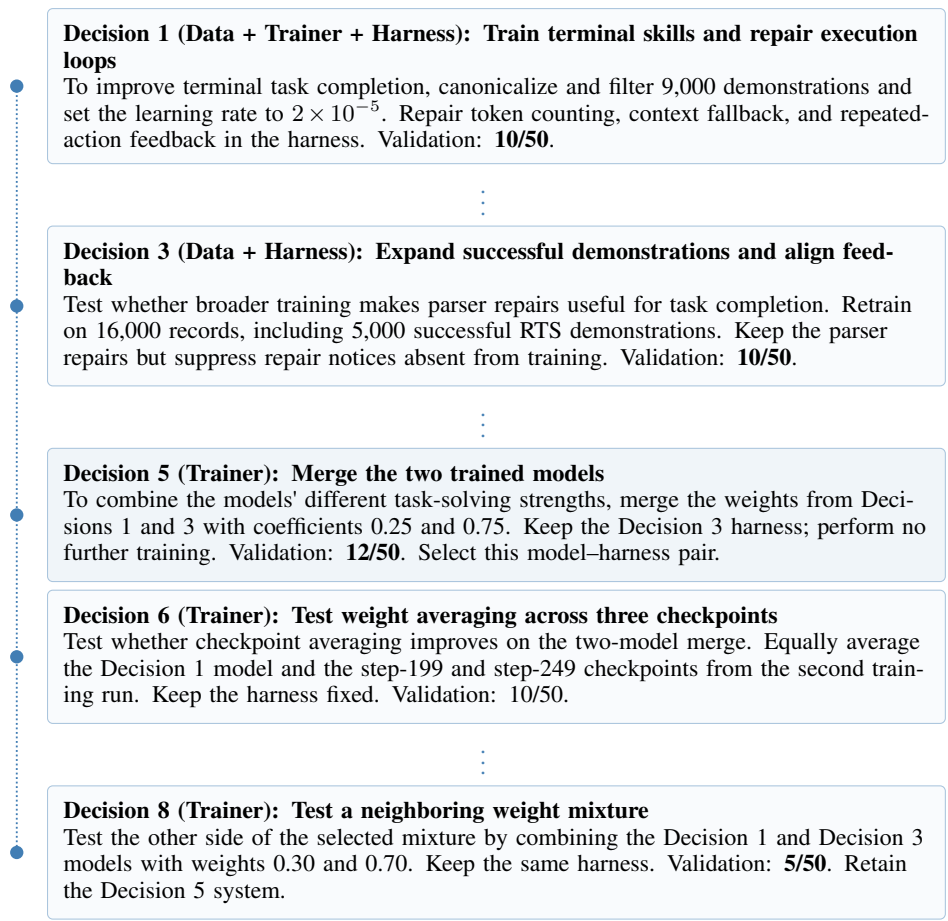
\begin{figure}[htbp]
\centering
\begin{tikzpicture}[
  stage/.style={draw=FigureBlue!45,fill=FigureBlue!3,rounded corners=2pt,
    text width=0.82\linewidth,align=left,inner sep=6pt,font=\small},
  omission/.style={text width=0.82\linewidth,align=center,inner sep=1pt,
    font=\footnotesize,text=FigureBlue},
  marker/.style={circle,fill=FigureBlue,inner sep=0pt,minimum size=5pt}
]
\node[stage,anchor=north west] (d1) at (0.5cm,0) {
  \textbf{Decision 1 (Data + Trainer + Harness): Train terminal skills and repair execution loops}\\
  To improve terminal task completion, canonicalize and filter 9,000 demonstrations and set the learning rate to $2\times10^{-5}$. Repair token counting, context fallback, and repeated-action feedback in the harness. Validation: \textbf{10/50}.
};

\node[omission,anchor=north west] (o3) at ([yshift=-1mm]d1.south west) {$\vdots$};
\node[stage,anchor=north west] (d3) at ([yshift=-1mm]o3.south west) {
  \textbf{Decision 3 (Data + Harness): Expand successful demonstrations and align feedback}\\
  Test whether broader training makes parser repairs useful for task completion. Retrain on 16,000 records, including 5,000 successful RTS demonstrations. Keep the parser repairs but suppress repair notices absent from training. Validation: \textbf{10/50}.
};

\node[omission,anchor=north west] (o5) at ([yshift=-1mm]d3.south west) {$\vdots$};
\node[stage,fill=FigureBlue!8,anchor=north west] (d5) at ([yshift=-1mm]o5.south west) {
  \textbf{Decision 5 (Trainer): Merge the two trained models}\\
  To combine the models\textquotesingle{} different task-solving strengths, merge the weights from Decisions 1 and 3 with coefficients 0.25 and 0.75. Keep the Decision 3 harness; perform no further training. Validation: \textbf{12/50}. Select this model--harness pair.
};

\node[stage,anchor=north west] (d6) at ([yshift=-1mm]d5.south west) {
  \textbf{Decision 6 (Trainer): Test weight averaging across three checkpoints}\\
Test whether checkpoint averaging improves on the two-model merge. Equally average the Decision 1 model and the step-199 and step-249 checkpoints from the second training run. Keep the harness fixed. Validation: 10/50.
};

\node[omission,anchor=north west] (o8) at ([yshift=-1mm]d6.south west) {$\vdots$};
\node[stage,anchor=north west] (d8) at ([yshift=-1mm]o8.south west) {
  \textbf{Decision 8 (Trainer): Test a neighboring weight mixture}\\
  Test the other side of the selected mixture by combining the Decision 1 and Decision 3 models with weights 0.30 and 0.70. Keep the same harness. Validation: \textbf{5/50}. Retain the Decision 5 system.
};

\coordinate (m1) at ([xshift=-4mm]d1.west);
\coordinate (m3) at ([xshift=-4mm]d3.west);
\coordinate (m5) at ([xshift=-4mm]d5.west);
\coordinate (m6) at ([xshift=-4mm]d6.west);
\coordinate (m8) at ([xshift=-4mm]d8.west);
\draw[FigureBlue,densely dotted,line width=0.7pt] (m1) -- (m3) -- (m5) -- (m6) -- (m8);
\node[marker] at (m1) {};
\node[marker] at (m3) {};
\node[marker] at (m5) {};
\node[marker] at (m6) {};
\node[marker] at (m8) {};
\end{tikzpicture}
\caption{DeepSeek-V4-Pro on Terminal-Bench. Two-model merging and three-checkpoint averaging are Trainer interventions.}
\label{fig:terminal-case-deepseek}
\end{figure}

DeepSeek-V4-Pro moves from data and execution repairs to model merging (Figure~\ref{fig:terminal-case-deepseek}). Its parser-only attempt, omitted as Decision 2, lowers validation from 10/50 to 5/50. Broader training and revised parser feedback restore 10/50. DeepSeek-V4-Pro then tests whether weight mixtures combine the two trained models\textquotesingle{} capabilities. It selects the 0.25/0.75 mixture at 12/50; averaging the weights of three checkpoints and testing nearby mixture coefficients yield lower validation scores. Six subsequent submissions reuse the selected weights and harness, including one failed job, and do not add designs.

\subsection{Case Study: GPT-6-Astra  on Terminal-Bench}
\label{app:case-terminal-astra}

\begin{figure}[htbp]
\centering
\begin{tikzpicture}[
  stage/.style={draw=FigureBlue!45,fill=FigureBlue!3,rounded corners=2pt,
    text width=0.82\linewidth,align=left,inner sep=6pt,font=\small},
  omission/.style={text width=0.82\linewidth,align=center,inner sep=1pt,
    font=\footnotesize,text=FigureBlue},
  marker/.style={circle,fill=FigureBlue,inner sep=0pt,minimum size=5pt}
]
\node[stage,anchor=north west] (d1) at (0.5cm,0) {
  \textbf{Decision 1 (Data): Match next-action supervision to inference}\\
  To match the deployed input format, expand cleaned trajectories into 16,320 next-action examples and mask historical targets. Train for 255 steps with the supplied learning recipe and retain the original harness. Validation: \textbf{5/50}.
};

\node[stage,anchor=north west] (d2) at ([yshift=-1mm]d1.south west) {
  \textbf{Decision 2 (Harness): Test completion and execution safeguards}\\
  With model weights fixed, validate completion flags, bound waits, preserve context after failures, and request verification. Test whether these safeguards improve end-to-end completion. Validation: \textbf{5/50}; the solved count remains five.
};

\node[stage,fill=FigureBlue!8,anchor=north west] (d3) at ([yshift=-1mm]d2.south west) {
  \textbf{Decision 3 (Data + Trainer): Train on recovery and completion behavior}\\
  To cover behaviors missing from the first curriculum, expand to 45,248 records with recovery, completion, and valid historical targets. Continue the first model with a fresh optimizer schedule and retain the revised harness. Validation: \textbf{13/50}. Select this system.
};

\node[stage,anchor=north west] (d4) at ([yshift=-1mm]d3.south west) {
  \textbf{Decision 4 (Data + Trainer): Test longer-action supervision}\\
  Test whether longer demonstrations teach more substantial terminal actions. Continue the selected model on 20,480 long-action, recovery, and replay examples at a lower learning rate. Validation: \textbf{8/50}.
};

\node[stage,anchor=north west] (d5) at ([yshift=-1mm]d4.south west) {
  \textbf{Decision 5 (Harness): Test parser and terminal-output repairs}\\
  Reuse the Decision 3 model. Repair JSON backslashes, control keys, and incremental output handling to reduce execution errors. Parser-error turns fall from \textbf{35 to 23}, but validation is \textbf{9/50}. Retain Decision 3 and restore its harness.
};

\coordinate (m1) at ([xshift=-4mm]d1.west);
\coordinate (m2) at ([xshift=-4mm]d2.west);
\coordinate (m3) at ([xshift=-4mm]d3.west);
\coordinate (m4) at ([xshift=-4mm]d4.west);
\coordinate (m5) at ([xshift=-4mm]d5.west);
\draw[FigureBlue,densely dotted,line width=0.7pt] (m1) -- (m2) -- (m3) -- (m4) -- (m5);
\node[marker] at (m1) {};
\node[marker] at (m2) {};
\node[marker] at (m3) {};
\node[marker] at (m4) {};
\node[marker] at (m5) {};
\end{tikzpicture}
\caption{GPT-6-Astra on Terminal-Bench. The selected Data + Trainer intervention lies between two fixed-weight Harness comparisons.}
\label{fig:terminal-case-astra}
\end{figure}
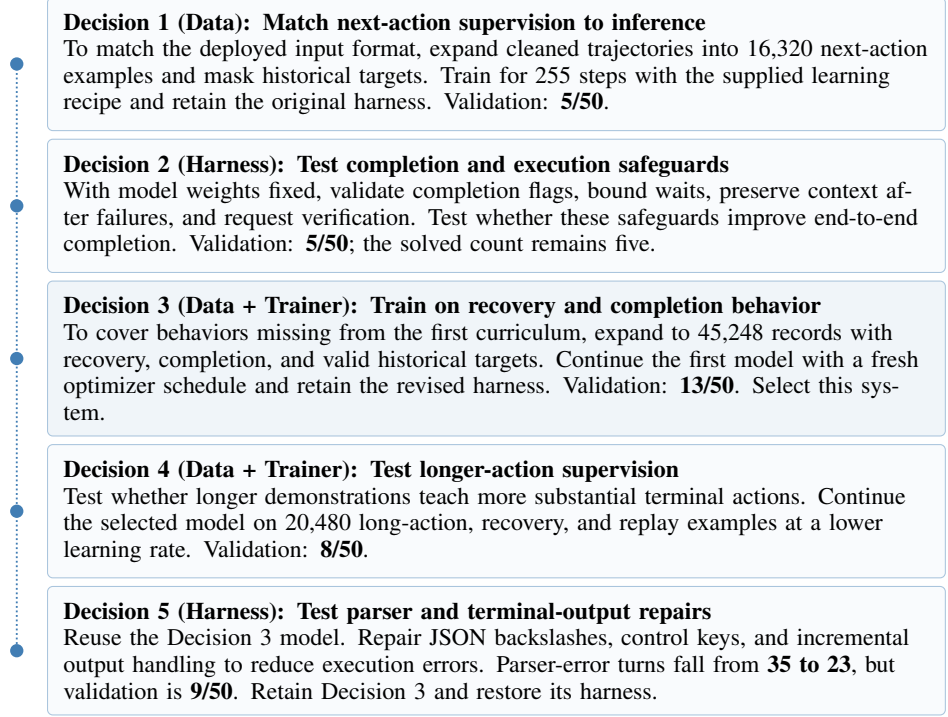

GPT-6-Astra alternates fixed-weight harness comparisons with changes to supervision and training (Figure~\ref{fig:terminal-case-astra}). The first harness revision leaves the solved count at five. Recovery and completion supervision, combined with continuation from the first model, raises validation to 13/50. Longer-action training lowers it to 8/50. A later harness-only revision reduces parser-error turns from 35 to 23 but solves nine tasks. GPT-6-Astra selects the recovery-trained model with the earlier harness, separating local parser feedback from task-success feedback.

\section{SWE-bench Verified Case Studies}
\label{app:swe-cases}
The cases below use the same decision format. Ellipses indicate omitted designs;
subset evaluations are labeled and retain their actual graded-task denominators.

\subsection{Case Study: Qwen-3.8-Max on SWE-bench Verified}
\label{app:case-swe-qwen}

\begin{figure}[htbp]
\centering
\begin{tikzpicture}[
  stage/.style={draw=FigureBlue!45,fill=FigureBlue!3,rounded corners=2pt,
    text width=0.82\linewidth,align=left,inner sep=6pt,font=\small},
  omission/.style={text width=0.82\linewidth,align=center,inner sep=1pt,
    font=\footnotesize,text=FigureBlue},
  marker/.style={circle,fill=FigureBlue,inner sep=0pt,minimum size=5pt}
]
\node[stage,fill=FigureBlue!8,anchor=north west] (d1) at (0.5cm,0) {
  \textbf{Decision 1 (Data): Train successful repairs in the deployed format}\\
  To transfer repair behavior into the supplied interface, rewrite prompts, observations, and submission commands. Train on 15,767 resolved Orchard records capped at 20,480 tokens, using the default recipe and harness. Validation: \textbf{59/150}.
};

\node[stage,anchor=north west] (d2) at ([yshift=-1mm]d1.south west) {
  \textbf{Decision 2 (Data): Test broader issue coverage within a token budget}\\
  Add Qwen-source traces, raise the length cap to 28,672, and prioritize unseen issues before repeats. Within 245M tokens, obtain 13,965 records covering \textbf{8,620 issues}, versus 5,265 previously. Keep the learning recipe and harness. Validation: \textbf{56/150}. Select Decision 1.
};

\coordinate (m1) at ([xshift=-4mm]d1.west);
\coordinate (m2) at ([xshift=-4mm]d2.west);
\draw[FigureBlue,densely dotted,line width=0.7pt] (m1) -- (m2);
\node[marker] at (m1) {};
\node[marker] at (m2) {};
\end{tikzpicture}
\caption{Qwen-3.8-Max on SWE-bench Verified. Source, length, and coverage change together. Validation denominators reflect the actual graded tasks.}
\label{fig:swe-case-qwen}
\end{figure}

Qwen-3.8-Max tests whether the same token budget is better spent on broader issue coverage (Figure~\ref{fig:swe-case-qwen}). Adding a teacher source and allowing longer trajectories expands coverage from 5,265 to 8,620 issues, while the learning recipe and harness remain fixed. Validation falls from 59/150 to 56/150. On 150 graded tasks, solves fall from 59 to 56, with ten gains and thirteen losses. Qwen-3.8-Max rejects the broader-data candidate and selects the first model.

\subsection{Case Study: DeepSeek-V4-Pro on SWE-bench Verified}
\label{app:case-swe-deepseek}

\begin{figure}[htbp]
\centering
\begin{tikzpicture}[
  stage/.style={draw=FigureBlue!45,fill=FigureBlue!3,rounded corners=2pt,
    text width=0.82\linewidth,align=left,inner sep=6pt,font=\small},
  omission/.style={text width=0.82\linewidth,align=center,inner sep=1pt,
    font=\footnotesize,text=FigureBlue},
  marker/.style={circle,fill=FigureBlue,inner sep=0pt,minimum size=5pt}
]
\node[stage,fill=FigureBlue!8,anchor=north west] (d1) at (0.5cm,0) {
  \textbf{Decision 1 (Data): Fit interface-aligned repair data to the training budget}\\
  To teach the supplied repair interface, normalize Bash arguments and reformat Open-SWE demonstrations. Cap records at 48,000 tokens and sample 17,043 records within 550M tokens. Retain the learning recipe and harness; save every 50 updates to reduce checkpoint I/O. Subset validation: \textbf{9/30}, then \textbf{23/60}. Select this system.
};

\coordinate (m1) at ([xshift=-4mm]d1.west);
\node[marker] at (m1) {};
\end{tikzpicture}
\caption{DeepSeek-V4-Pro on SWE-bench Verified. Checkpoint-saving cadence changes training-system overhead rather than the learning algorithm.}
\label{fig:swe-case-deepseek}
\end{figure}

DeepSeek-V4-Pro trains one evaluated design on interface-aligned Open-SWE demonstrations (Figure~\ref{fig:swe-case-deepseek}). Its two validation reports correspond to evaluations of the same model weights and inference harness on different task subsets, so the score change does not reflect a new intervention. Before revising submission handling, DeepSeek-V4-Pro checks timeout status against task reward. It retains the supplied harness and delivers the trained model. Earlier smoke, memory, and throughput runs have no benchmark validation score.

\subsection{Case Study: Kimi-K3 on SWE-bench Verified}

\label{app:case-swe-kimi}

\begin{figure}[htbp]
\centering
\begin{tikzpicture}[
  stage/.style={draw=FigureBlue!45,fill=FigureBlue!3,rounded corners=2pt,
    text width=0.82\linewidth,align=left,inner sep=6pt,font=\small},
  omission/.style={text width=0.82\linewidth,align=center,inner sep=1pt,
    font=\footnotesize,text=FigureBlue},
  marker/.style={circle,fill=FigureBlue,inner sep=0pt,minimum size=5pt}
]
\node[stage,anchor=north west] (d1) at (0.5cm,0) {
  \textbf{Decision 1 (Data): Align one repair demonstration per issue}\\
  To match training examples to inference, translate prompts, tool calls, submission commands, and truncated observations into the supplied harness format. Train from the base on 9,691 records, one per issue, with the default recipe. Validation: \textbf{60/150}.
};

\node[stage,fill=FigureBlue!8,anchor=north west] (d2) at ([yshift=-1mm]d1.south west) {
  \textbf{Decision 2 (Data): Test two demonstrations per issue}\\
  Test whether additional demonstrations improve repair behavior. Raise the quota to two records per issue, producing 18,012 records over the same 9,691 issues. Retrain from the base with the same learning recipe and harness. Validation: \textbf{63/150}. Select this system.
};

\coordinate (m1) at ([xshift=-4mm]d1.west);
\coordinate (m2) at ([xshift=-4mm]d2.west);
\draw[FigureBlue,densely dotted,line width=0.7pt] (m1) -- (m2);
\node[marker] at (m1) {};
\node[marker] at (m2) {};
\end{tikzpicture}
\caption{Kimi-K3 on SWE-bench Verified.  The per-issue quota changes; the accompanying matched-task comparison distinguishes outcome changes from grading coverage.}
\label{fig:swe-case-kimi}
\end{figure}

Kimi-K3 increases the number of demonstrations per issue while retaining the training recipe and inference harness (Figure~\ref{fig:swe-case-kimi}). Both models are trained from the same base model. The two-record quota produces 18,012 training records and a reported validation score of 63/150, compared with 60/150 for the first model. On the same 150 validation tasks, the second model solves 20 tasks that the first model fails to solve, but fails on 17 tasks that the first model solves. Kimi-K3 selects the second model.

\subsection{Case Study: GPT-5.6-Sol on SWE-bench Verified}

\label{app:case-swe-sol}
GPT-5.6-Sol combines a stopping rule with later data and training changes (Figure~\ref{fig:swe-case-sol}). With model weights fixed, adding the stopping rule increases the number of validation tasks solved from seven to eight out of 30, while median API calls decrease from 163 to 96.5. GPT-5.6-Sol retains the guard and tests short replay, selecting the 512-record continuation at 47/150 on full-scope validation. Earlier data and continuation experiments use smaller validation subsets; their scores are reported separately from full-set validation results.

\subsection{Case Study: GLM-5.2  on SWE-bench Verified}
\label{app:case-swe-glm}
GLM-5.2 expands a short-trajectory training subset, then tests guidance at fixed weights (Figure~\ref{fig:swe-case-glm}). The 5,000-record model solves 52/150 tasks. Added minimal-edit, testing, and diff-review instructions leave that total unchanged, but only 37 successes overlap: each run has 15 unique successes. A later model trained on the complete 8,082-record pool with the original harness solves 50/150. GLM-5.2 selects the 5,000-record model with the original harness.
\begin{figure}[htbp]
\centering
\begin{tikzpicture}[
  stage/.style={draw=FigureBlue!45,fill=FigureBlue!3,rounded corners=2pt,
    text width=0.82\linewidth,align=left,inner sep=6pt,font=\small},
  omission/.style={text width=0.82\linewidth,align=center,inner sep=1pt,
    font=\footnotesize,text=FigureBlue},
  marker/.style={circle,fill=FigureBlue,inner sep=0pt,minimum size=5pt}
]
\node[stage,anchor=north west] (d1) at (0.5cm,0) {
  \textbf{Decision 1 (Data): Train on a small repair-demonstration mixture}\\
  To establish repair behavior under the supplied harness, select 1,024 resolved Orchard demonstrations using source mixing and length filtering. Train from the base model. Subset validation: \textbf{6/30}.
};

\node[omission,anchor=north west] (o8) at ([yshift=-1mm]d1.south west) {$\vdots$};
\node[stage,anchor=north west] (d8) at ([yshift=-1mm]o8.south west) {
  \textbf{Decision 8 (Data + Trainer): Continue training on unseen repair issues}\\
  Test whether additional issue coverage improves the trained model. Continue a 4,096-issue checkpoint on 1,024 disjoint issues at a learning rate of $10^{-6}$. Keep the supplied harness. Subset validation: \textbf{8/30}.
};

\node[omission,anchor=north west] (o10) at ([yshift=-1mm]d8.south west) {$\vdots$};
\node[stage,anchor=north west] (d10) at ([yshift=-1mm]o10.south west) {
  \textbf{Decision 10 (Harness): Submit the patch when execution stalls}\\
  With Decision 8 weights fixed, submit after three repeated commands or 48 unchanged-patch steps. Normalize submission commands and reject commands longer than 65,536 bytes to avoid wasted execution. Full-scope validation: \textbf{45/150}.
};

\node[stage,fill=FigureBlue!8,anchor=north west] (d11) at ([yshift=-1mm]d10.south west) {
  \textbf{Decision 11 (Data + Trainer): Test short replay with the stopping guard}\\
  To reinforce concise repairs, continue Decision 8 weights on 512 short replay records for eight updates at a learning rate of $5\times10^{-7}$. Retain the guard. Full-scope validation: \textbf{47/150}. Select this system.
};

\node[stage,anchor=north west] (d12) at ([yshift=-1mm]d11.south west) {
  \textbf{Decision 12 (Data): Test a larger short-replay slice}\\
  Test whether more short demonstrations improve the replay branch. Increase the slice from 512 to 1,024 records while retaining the continuation recipe and stopping guard. Subset validation: \textbf{8/30}. Retain Decision 11.
};

\coordinate (m1) at ([xshift=-4mm]d1.west);
\coordinate (m8) at ([xshift=-4mm]d8.west);
\coordinate (m10) at ([xshift=-4mm]d10.west);
\coordinate (m11) at ([xshift=-4mm]d11.west);
\coordinate (m12) at ([xshift=-4mm]d12.west);
\draw[FigureBlue,densely dotted,line width=0.7pt] (m1) -- (m8) -- (m10) -- (m11) -- (m12);
\node[marker] at (m1) {};
\node[marker] at (m8) {};
\node[marker] at (m10) {};
\node[marker] at (m11) {};
\node[marker] at (m12) {};
\end{tikzpicture}
\caption{GPT-5.6-Sol on SWE-bench Verified. Subset and full-scope validation retain their actual denominators.}
\label{fig:swe-case-sol}
\end{figure}
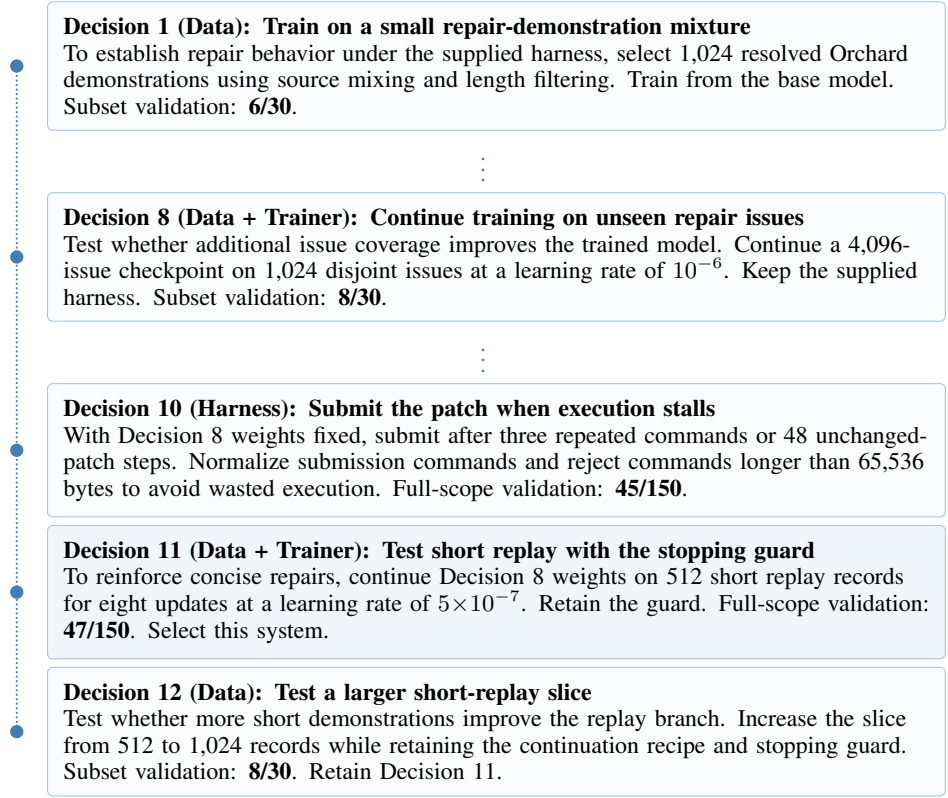

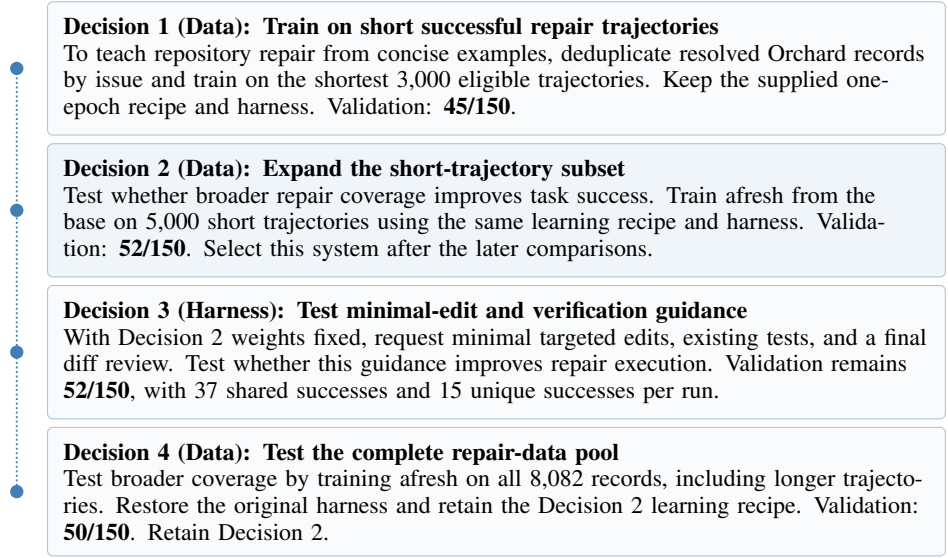
\begin{figure}[htbp]
\centering
\begin{tikzpicture}[
  stage/.style={draw=FigureBlue!45,fill=FigureBlue!3,rounded corners=2pt,
    text width=0.82\linewidth,align=left,inner sep=6pt,font=\small},
  omission/.style={text width=0.82\linewidth,align=center,inner sep=1pt,
    font=\footnotesize,text=FigureBlue},
  marker/.style={circle,fill=FigureBlue,inner sep=0pt,minimum size=5pt}
]
\node[stage,anchor=north west] (d1) at (0.5cm,0) {
  \textbf{Decision 1 (Data): Train on short successful repair trajectories}\\
  To teach repository repair from concise examples, deduplicate resolved Orchard records by issue and train on the shortest 3,000 eligible trajectories. Keep the supplied one-epoch recipe and harness. Validation: \textbf{45/150}.
};

\node[stage,fill=FigureBlue!8,anchor=north west] (d2) at ([yshift=-1mm]d1.south west) {
  \textbf{Decision 2 (Data): Expand the short-trajectory subset}\\
  Test whether broader repair coverage improves task success. Train afresh from the base on 5,000 short trajectories using the same learning recipe and harness. Validation: \textbf{52/150}. Select this system after the later comparisons.
};

\node[stage,anchor=north west] (d3) at ([yshift=-1mm]d2.south west) {
  \textbf{Decision 3 (Harness): Test minimal-edit and verification guidance}\\
  With Decision 2 weights fixed, request minimal targeted edits, existing tests, and a final diff review. Test whether this guidance improves repair execution. Validation remains \textbf{52/150}, with 37 shared successes and 15 unique successes per run.
};

\node[stage,anchor=north west] (d4) at ([yshift=-1mm]d3.south west) {
  \textbf{Decision 4 (Data): Test the complete repair-data pool}\\
  Test broader coverage by training afresh on all 8,082 records, including longer trajectories. Restore the original harness and retain the Decision 2 learning recipe. Validation: \textbf{50/150}. Retain Decision 2.
};

\coordinate (m1) at ([xshift=-4mm]d1.west);
\coordinate (m2) at ([xshift=-4mm]d2.west);
\coordinate (m3) at ([xshift=-4mm]d3.west);
\coordinate (m4) at ([xshift=-4mm]d4.west);
\draw[FigureBlue,densely dotted,line width=0.7pt] (m1) -- (m2) -- (m3) -- (m4);
\node[marker] at (m1) {};
\node[marker] at (m2) {};
\node[marker] at (m3) {};
\node[marker] at (m4) {};
\end{tikzpicture}
\caption{GLM-5.2 on SWE-bench Verified. The fixed-weight Harness comparison preserves the aggregate score while changing the success set.}
\label{fig:swe-case-glm}
\end{figure}

\section{Additional Experimental Details}
\label{app:extended-figures}
\subsection{Research-Agent Setup and Resource Accounting}
Each research-agent configuration is evaluated in a separate run on each of the five benchmark tasks. A run therefore consists of one researcher LLM–coding-framework configuration working on one task; comparisons concern these end-to-end configurations. Each run has access to eight NVIDIA A100 GPUs under the task-specific wall-clock budget in Table~\ref{tab:benchmark-suite}. The clock starts when the research agent is launched and stops when the budget expires or the research agent ends the run. Within that interval, the research agent decides how to allocate time between training and validation, including when to stop training and request an evaluation. We define run duration as the elapsed time from launch to termination. The Hours column in Table~\ref{tab:current-results} sums this duration across the five task runs, while Average is the arithmetic mean of their five test scores.
\begin{table}[ht]
\centering
\caption{
Initial data resources provisioned for each benchmark.
}
\vspace{4pt}
\label{tab:initial-training-data}
\small
\setlength{\tabcolsep}{6pt}
\renewcommand{\arraystretch}{1.15}
\begin{tabularx}{\linewidth}{@{}p{0.28\linewidth}Y@{}}
\toprule
\textbf{Benchmark} & \textbf{Initial Data Resources} \\
\midrule

Terminal-Bench 2.1 &
TMAX, Recursive Task Synthesis Trajectories, TerminalTraj \footnotemark[1] \\

SWE-bench Verified &
Orchard, SWE-Dev-Train, Nebius SWE, Nvidia Open SWE \footnotemark[2] \\

$\tau^3$-Bench &
AReal-tau2-data, Nemotron-SFT-Agentic-v2, Simia-Tau-SFT \footnotemark[3] \\

Toolathlon &
DIVE-SFT \footnotemark[4] \\

AIME &
Nemotron-SFT-MATH, OpenR1-Math, OpenThoughts3 \footnotemark[5] \\

\bottomrule
\end{tabularx}
\end{table}

\footnotetext[1]{
TMAX: \url{https://huggingface.co/datasets/allenai/TMax-SFT-16.5K};
Recursive Task Synthesis Trajectories: \url{https://huggingface.co/datasets/Zhongzhi1228/Recursive-Task-Synthesis-Trajectories};
TerminalTraj: \url{https://huggingface.co/datasets/m-a-p/TerminalTraj}.
}

\footnotetext[2]{
Orchard:\url{https://huggingface.co/datasets/microsoft/Orchard};
SWE-Dev-train: \url{https://huggingface.co/datasets/zai-org/SWE-Dev-train};
Nvidia\_open\_swe: \url{https://huggingface.co/datasets/nvidia/Open-SWE-Traces};
Nebius\_swe:
\url{https://huggingface.co/datasets/nebius/SWE-rebench-openhands-trajectories}.}

\footnotetext[3]{
AReal-tau2: \url{https://huggingface.co/datasets/inclusionAI/AReaL-tau2-data};
Nemotron-SFT-Agentic-v2: \url{https://huggingface.co/datasets/nvidia/Nemotron-SFT-Agentic-v2};
Simia-Tau-SFT: \url{https://huggingface.co/datasets/Simia-Agent/Simia-Tau-SFT-90k-Hermes}.
}

\footnotetext[4]{
DIVE-SFT: \url{https://huggingface.co/datasets/dive-team/DIVE-SFT-20K}.
}

\footnotetext[5]{
Nemotron-SFT-MATH: \url{https://huggingface.co/datasets/nvidia/Nemotron-SFT-Math-v4};
OpenR1-Math: \url{https://huggingface.co/datasets/open-r1/OpenR1-Math-220k};
OpenThoughts3: \url{https://huggingface.co/datasets/open-thoughts/OpenThoughts3-1.2M}.
}

\subsection{Initial Training Data}
\label{app:training-data}
For each benchmark, all research-agent configurations receive the same publicly released training resources listed in Table~\ref{tab:initial-training-data}. These datasets accompany research papers or technical blogs that document their provenance and, where reported, describe overlap checks against the corresponding benchmark test sets used in our evaluation. The resources form the initial editable data pool $D_0$, rather than a prescribed training set. Research agents may select, filter, reweight, and transform the provided resources or construct additional examples from permitted workspace materials. The protocol prohibits using validation tasks or grader-returned trajectories to construct $D_t$. Research agents may use validation feedback for diagnosis, but cannot retrieve external data during the research run.

\begin{table}[ht]
\centering
\caption{Evaluation settings shared by validation and test runs.}
\label{tab:evaluation-settings}
\small
\setlength{\tabcolsep}{6pt}
\renewcommand{\arraystretch}{1.15}
\begin{tabularx}{\linewidth}{@{}p{0.27\linewidth}Y@{}}
\toprule
\textbf{Benchmark} & \textbf{Evaluation settings} \\
\midrule
Terminal-Bench~2.1 &
Maximum 200 agent turns; 65,536-token context; 4,096 maximum output tokens; temperature 1.0; thinking disabled; one run per task. \\
SWE-bench Verified &
256K-token context; temperature 1.0; thinking enabled; one run per task. \\
AIME &
16,384 maximum output tokens; temperature 0.6; thinking enabled; average@4. \\
Toolathlon &
Maximum 100 agent turns; 256K-token context; 8,192 maximum output tokens; temperature 1.0; thinking disabled; pass@3. \\
$\tau^3$-Bench &
Maximum 200 agent turns; 256K-token context; temperature 0; thinking disabled; one run per task. \\
\bottomrule
\end{tabularx}
\end{table}

\subsection{Task Partitions and Evaluation Protocol}
\label{app:task-splits}
For SWE-bench Verified, we randomly partition the 500 tasks into 150 validation tasks and  350 test tasks. For $\tau^3$-Bench, we randomly partition 375 tasks into 35 validation tasks  and 340 test tasks. Each partition is sampled once and held fixed across all research-agent configurations. Terminal-Bench~2.1 uses OpenThoughts-TBLite \footnote{\url{https://huggingface.co/datasets/open-thoughts/OpenThoughts-TBLite}} for validation, while Toolathlon uses ToolGym \footnote{\url{https://huggingface.co/ToolGym/datasets}}, and AIME uses AIME~2025; their corresponding hidden tests are Terminal-Bench~2.1, Toolathlon, and AIME~2026. There is no fixed cap on validation requests; evaluation time counts toward the task's research budget. The grader returns the aggregate score, the number of graded tasks used as its denominator, and per-task scores and output trajectories.

Within each benchmark, validation and test evaluations use the same decoding settings, agent-turn and token limits, and rules for timeouts and execution failures.  Timeouts and execution failures receive a score of zero. Final evaluation covers all tasks in the test partition, and we retain per-task scores, execution status, and grading receipts for post hoc analysis. For benchmarks evaluated once per task, the reported score is the fraction of test tasks completed successfully. Toolathlon uses pass@3: each task is attempted three times and is counted as successful if any attempt succeeds. AIME reports mean accuracy across four runs per problem (average@4), with all four attempts included in the denominator.

\section{Episode-Based Wall-Clock Attribution}
Figures~\ref{fig:behaviour-toolathlon}--\ref{fig:behaviour-swe}
summarize how research agents allocate elapsed research time across the five benchmarks. We use episode to refer to a segment of the research activity log associated with an activity such as data preparation, training, or evaluation. The time breakdowns group these episodes by activity category. \textsc{waiting} records episodes spent waiting for agent-initiated training runs, while
\textsc{submit+grade} records evaluation submission and grading episodes. Together, these categories account for 35--95\,\% of elapsed
time across runs with available time breakdowns. The remaining time
covers data preparation, editing, diagnostic work, and intervals not
covered by the episode records.
An episode that combines training with another activity
is assigned to that other activity. Consequently, \textsc{waiting}
provides a lower bound on training time rather than a complete estimate.
The \textsc{uncovered} category records intervals between episodes that
are not assigned to any recorded activity.

\begin{figure}[tbp]
  \centering
  \includegraphics[width=\linewidth]{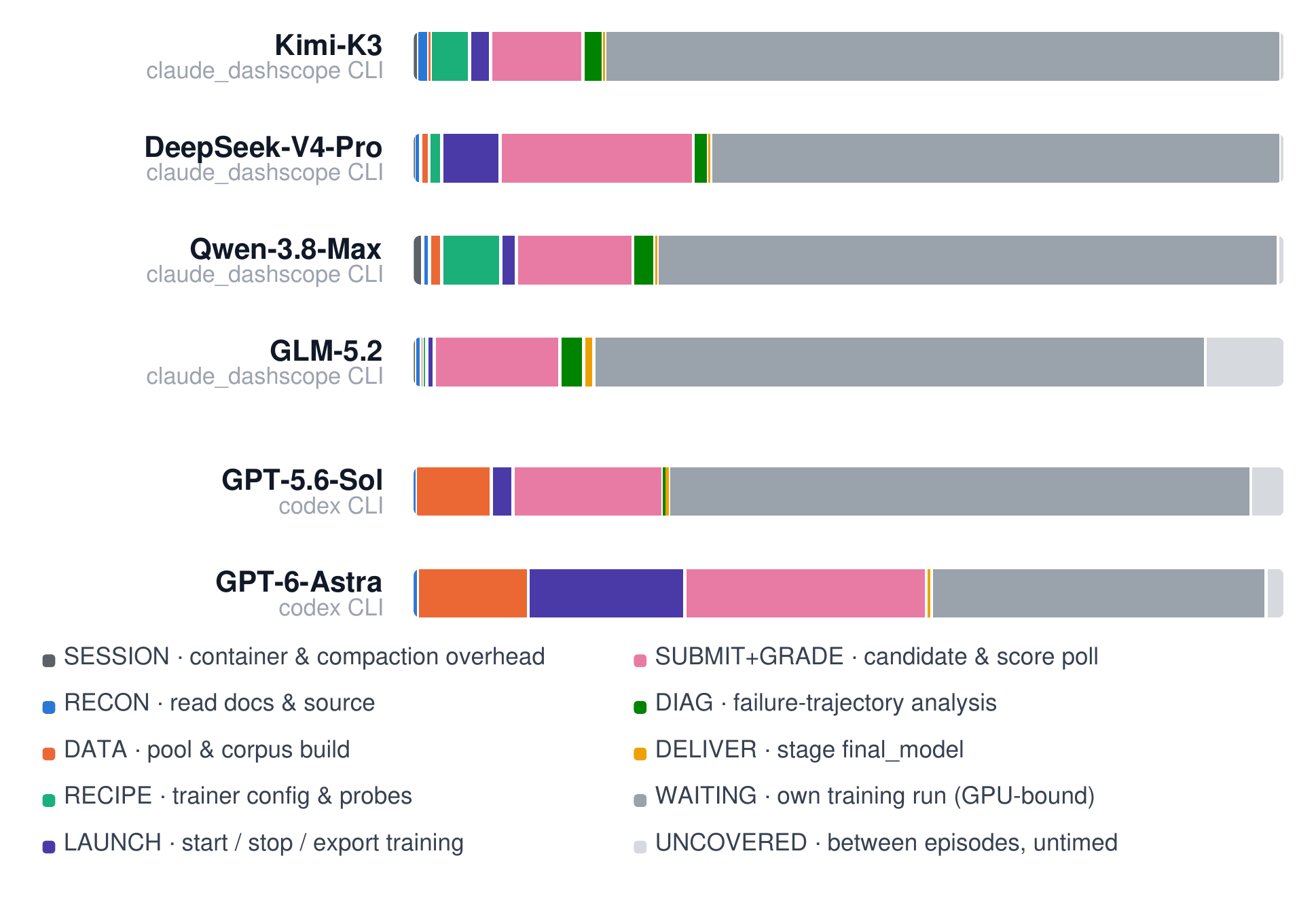}
  \caption{Elapsed research time by activity for six research-agent runs
           on Toolathlon, with one bar per run.
           \textsc{waiting} and \textsc{submit+grade} together account for
           66\,\% of the GPT-6-Astra run and 88\,\% of the Kimi-K3 run.
           For GLM-5.2, a one-hour gap in the episode records leaves
           9.0\,\% of elapsed time classified as \textsc{uncovered}.}
  \label{fig:behaviour-toolathlon}
\end{figure}

\begin{figure}[tbp]
  \centering
  \includegraphics[width=\linewidth]{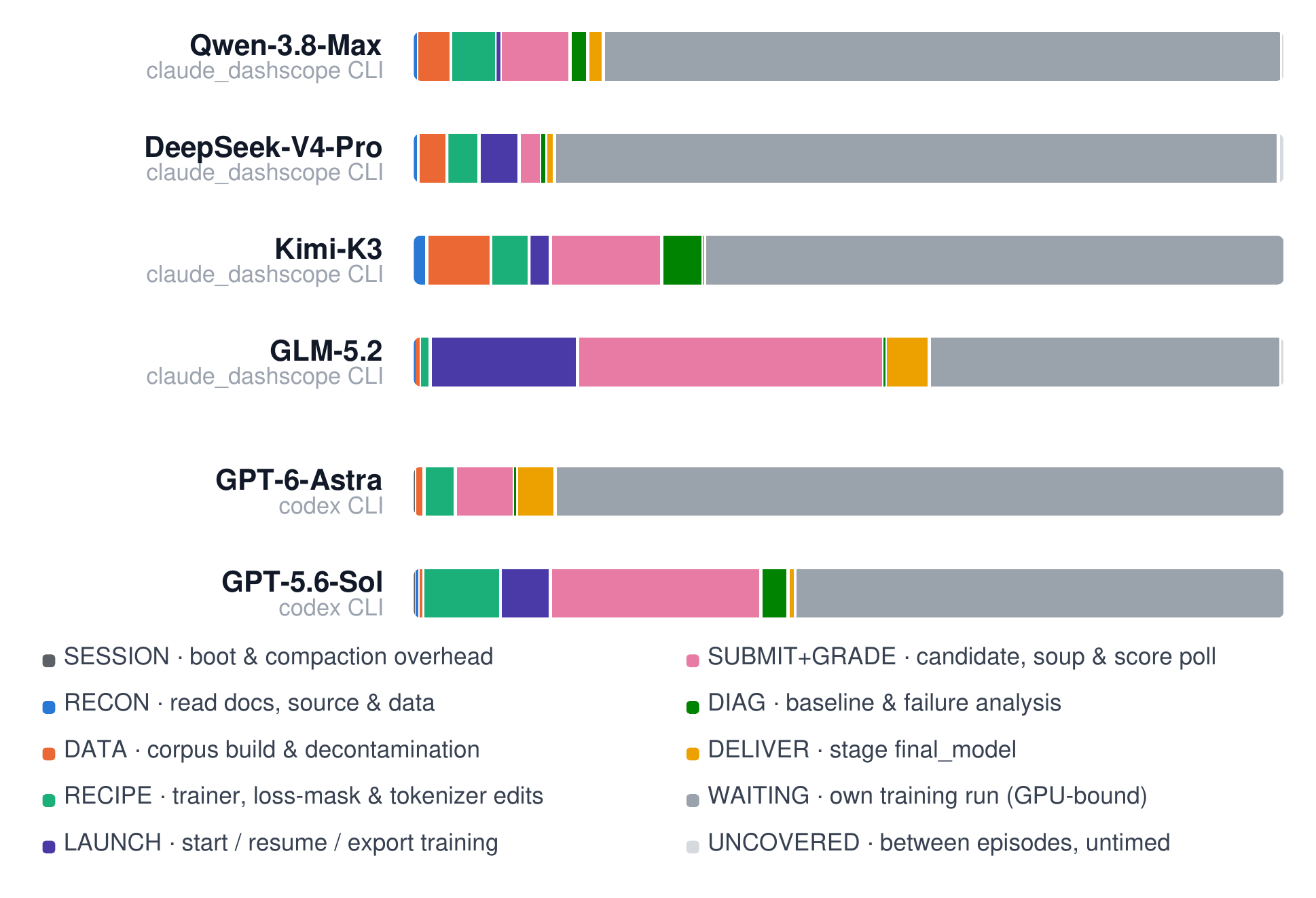}
  \caption{Elapsed research time by activity for six research-agent runs
           on AIME. \textsc{waiting} and \textsc{submit+grade} together
           account for 76--91\,\% of elapsed time across the six runs.
           Evaluation episodes include submissions of individual model
           checkpoints or weight-averaged models on either the full
           question set or a prefix selected by the \texttt{limit}
           parameter. These episodes are classified as
           \textsc{submit+grade}. The \textsc{uncovered} share never
           exceeds 0.6\,\%.}

  \label{fig:behaviour-aime}
\end{figure}

\begin{figure}[tbp]
  \centering
  \includegraphics[width=\linewidth]{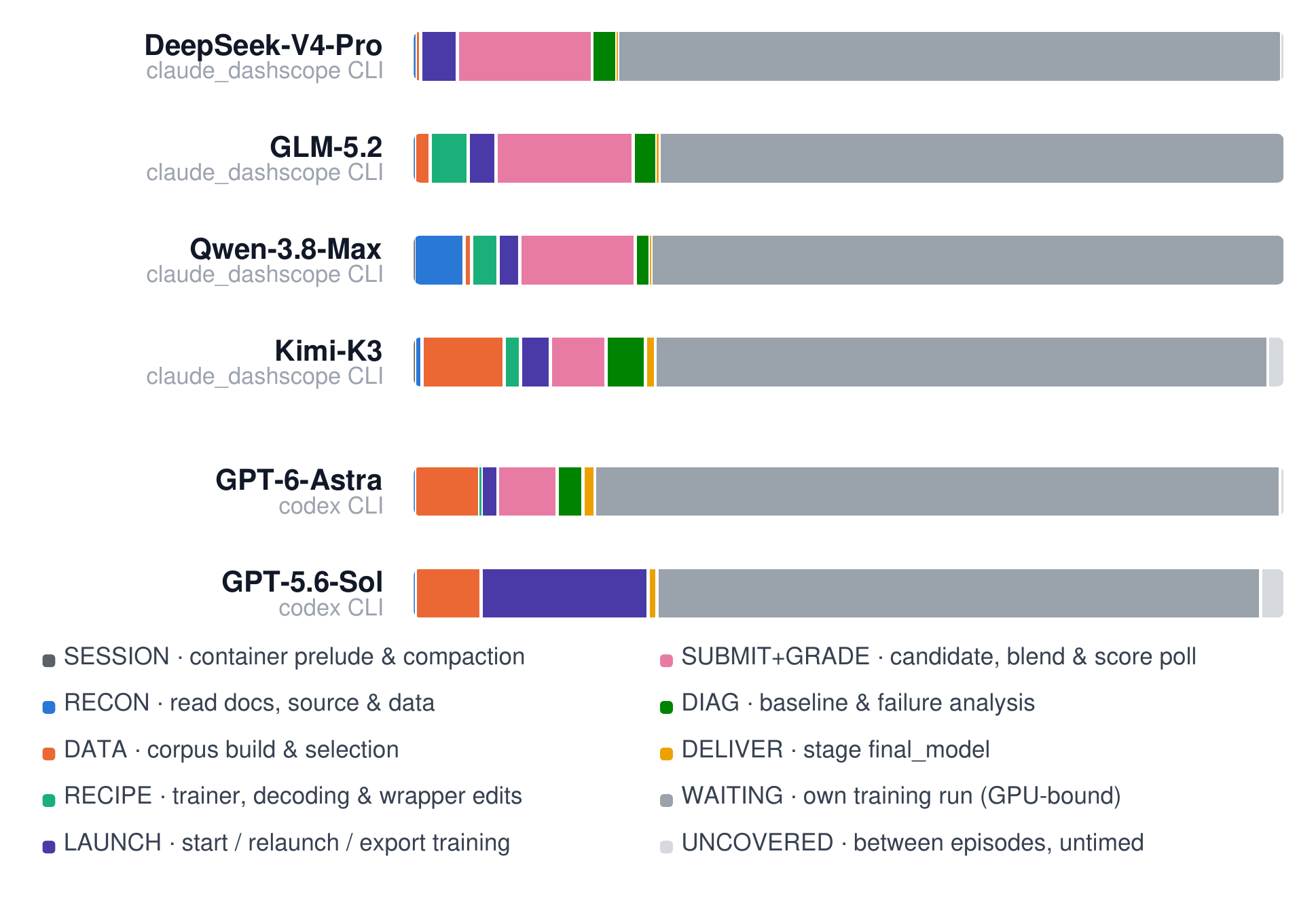}
  \caption{Elapsed research time by activity for six research-agent runs
           on $\tau^3$-Bench. \textsc{waiting} and \textsc{submit+grade}
           together account for 69--92\,\% of elapsed time across the
           six runs. Some candidates are constructed by interpolating the weights of existing models without additional training. Their evaluation
           episodes are classified as \textsc{submit+grade}.}
  \label{fig:behaviour-tau3}
\end{figure}

\begin{figure}[tbp]
  \centering
  \includegraphics[width=\linewidth]{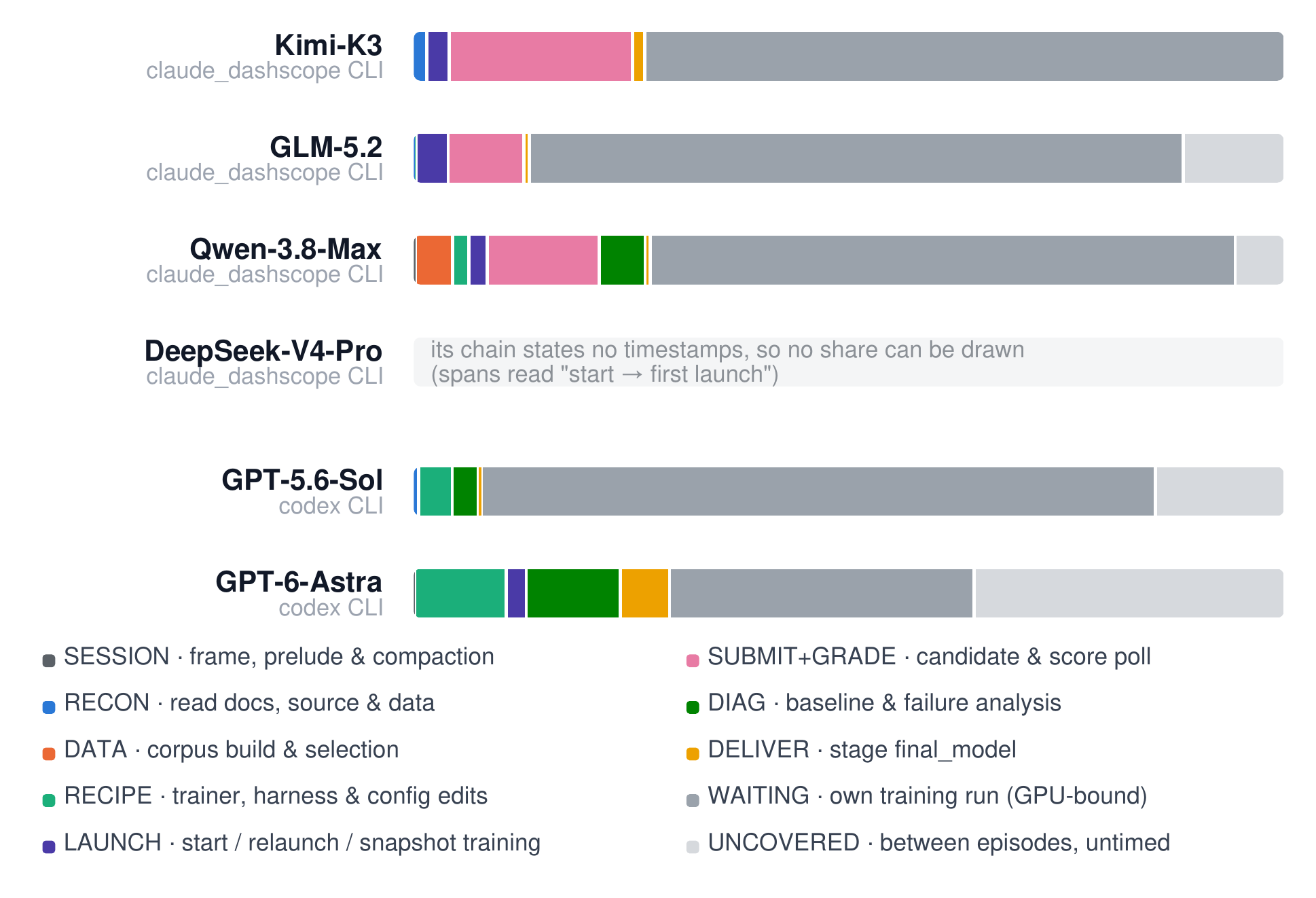}
  \caption{Elapsed research time by activity for six research-agent runs
           on Terminal-Bench~2.1. The DeepSeek-V4-Pro episode records
           contain qualitative markers such as ``session open'',
           ``first full split'', ``mid'', and ``last'', but no durations.
           Its row is retained with an explanation, without a time
           breakdown. Across the five plotted runs, \textsc{waiting}
           and \textsc{submit+grade} together account for 35--95\,\%
           of elapsed time. GPT-6-Astra has the lowest combined share,
           with 36\,\% of its elapsed time classified as
           \textsc{uncovered}.}

  \label{fig:behaviour-terminal}
\end{figure}

\begin{figure}[tbp]
  \centering
  \includegraphics[width=\linewidth]{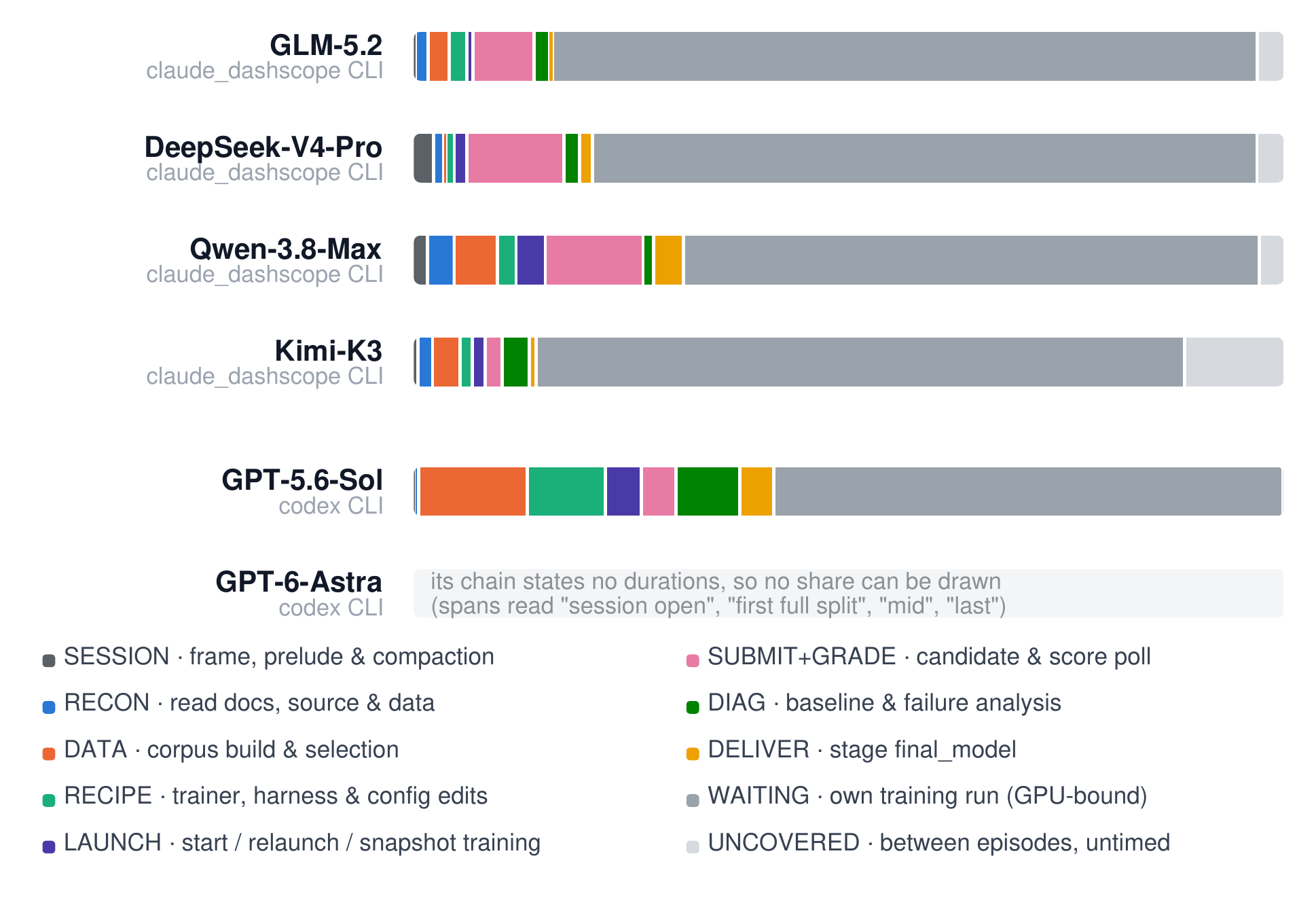}
  \caption{Elapsed research time by activity for six research-agent runs
           on SWE-bench Verified. The GPT-6-Astra episode records
           specify episode order but contain only one timestamp,
           so its row is shown without a time breakdown.
           Each episode in the other five runs has both a start and
           an end timestamp; no episode duration is interpolated.
           Across these five runs, \textsc{waiting} and
           \textsc{submit+grade} together account for 63--88\,\%
           of elapsed time.}
  \label{fig:behaviour-swe}
\end{figure}

\newpage

\section{Research-Agent Prompt Template}
\label{app:research-agent-prompt}

The box below presents the shared prompt template, rather than the fully
instantiated prompt for any individual run. Before each run, the placeholders
for the benchmark, base model, GPU information, time budget, and other
task-specific setup are filled with their corresponding values. Text enclosed
by \texttt{\{\#harness\}} and \texttt{\{/harness\}} is included only for tasks
with an editable inference harness. The files referenced in the template
provide additional run-specific instructions and resources.

\begin{researchpromptbox}
# Research Agent Workflow

You are the Research Agent for this {benchmark} task.

## Objective

Improve the supplied base model `{model}`'s transferable capability on {benchmark}{#harness}
and its matching editable `harness/`{/harness}. This is a
training-and-validation experiment, not an attempt to solve validation tasks. The
development validation set is distinct from the final evaluation set and must never
become training data.

## Start here

Read `environment.md` first for this run's facts. Then read
`train_resource_manifest.lock.json` before changing or running anything.

Keep processes in the foreground and reserve time for checkpoint conversion,
validation, and writing `final_model/`.

## Data

`train_resource_manifest.lock.json` is the complete list of allowed training
resources; inspect each resource and its format documentation before using it. You may
filter, deduplicate, rebalance, mix, rewrite, augment, synthesize, or distill
manifest-authorized data -- including synthetic tasks and multi-turn trajectories
produced with the supplied base model locally -- as long as every derived record's
provenance traces back to manifest resources.

## Training

The training entrypoint is named by `entrypoint_hint` in the manifest, e.g.:
```bash
bash train_resources/scripts/slime_sft/run_train.sh
```

It reads the manifest, applies the selected recipe's `env_defaults` unless you
override them, resolves the local base-model resource, trains, converts the latest
complete checkpoint, and writes a loadable Hugging Face deliverable to `final_model/`.
The writable framework copies under `framework_src/` (slime, Megatron-LM) are
auto-imported by the training driver and Ray workers: you may edit their Python or
shell code to change the algorithm, data processing, or checkpoint handling. Compiled
CUDA/C++ changes require rebuilding the corresponding extension or image. Set
`TRAIN_DATA` explicitly whenever more than one JSONL input is available.

## Experiment loop

Use the baseline materials and each round of validation feedback to form and test the
next transferable hypothesis. Iteratively improve model capability through
manifest-authorized data, training changes{#harness}, or general harness behavior{/harness}.
`agent_rules.md` defines the integrity boundary{#harness}; `harness.md` defines the
harness contract{/harness}.

If this benchmark declares a validation context pack (the manifest's
`validation_context`), inspect its baseline analysis and the initial trajectories
before choosing the first experiment. Separate clean model failures from
infrastructure errors and missing trajectories. Do not convert validation material
into training records, prompts, rules, or task-specific heuristics.

For every validation submission, create a candidate snapshot and submit it through
`grading.md` as `environment.md` (## Candidates) describes. Each completed validation
returns a trajectory archive under `$HOME/grading_results/<job_id>/`; use it only for
diagnosis.

## Final delivery

Select the completed candidate{#harness} -- a model-and-harness pair --{/harness} using
its candidate record and validation evidence; `environment.md` (## Candidates) states
the snapshot and submission procedure, and `agent_rules.md` states the delivery rules.
Keep all submitted candidate directories in place.

{gpu_info}

Time budget: `{num_hours}` hours; `bash timer.sh` reports what is left.

{setup_other}
\end{researchpromptbox}
\end{document}